%% file: main.tex
\documentclass{article}

\usepackage[preprint]{neurips_2024}   

\usepackage[utf8]{inputenc} % allow utf-8 input
\usepackage[T1]{fontenc}    % use 8-bit T1 fonts
\usepackage{hyperref}       % hyperlinks
\usepackage{url}            % simple URL typesetting
\usepackage{booktabs}       % professional-quality tables
\usepackage{amsfonts}       % blackboard math symbols
\usepackage{nicefrac}       % compact symbols for 1/2, etc.
\usepackage{microtype}      % microtypography
\usepackage{xcolor}         % colors
\usepackage{graphicx} 
\usepackage{amsmath}
\usepackage{cleveref}
\usepackage{pifont}
\usepackage{multirow}
\usepackage{amssymb}
\usepackage{listings}
\usepackage{subcaption}
\usepackage{float}
\usepackage{authblk}
\usepackage{textcomp}
\usepackage{ragged2e}
\usepackage{float}
\usepackage{bbding}
\usepackage{amssymb}  
\usepackage[accsupp]{axessibility}
\usepackage{bbding}

\title{SCBench: A Sports Commentary Benchmark for Video LLMs}

\author{
Kuangzhi Ge\textsuperscript{\rm 1$^{*}$}, Lingjun Chen\textsuperscript{\rm 1$^{*}$}, Kevin Zhang\textsuperscript{\rm 1$^{\dagger}$}, Yulin Luo\textsuperscript{\rm 1}, Tianyu Shi\textsuperscript{\rm 2},\\ Liaoyuan Fan\textsuperscript{\rm 3}, Xiang Li\textsuperscript{\rm 4}, Guanqun Wang\textsuperscript{\rm 1}, Shanghang Zhang\textsuperscript{\rm 1}~\textsuperscript{\Envelope}
\vspace{0.3cm}\\
\textsuperscript{\rm 1} State Key Laboratory of Multimedia Information Processing, School of Computer Science,\\Peking University; \textsuperscript{\rm 2} Department of Computer Science, University of Toronto;\\ \textsuperscript{\rm 3} The University of Hong Kong; \textsuperscript{\rm 4}Southern University of Science and Technology; \\
\vspace{0.3cm}
$^{*}$ Equal contribution, $^{\dagger}$ Project lead, \Envelope Corresponding author\\
\vspace{0.2cm}
 {\tt\small \{gekuangzhi, clj031031, yulin\}@stu.pku.edu.cn \{kevinzyz, wgq, shanghang\}@pku.edu.cn} \\ tianyu.s@outlook.com u3619617@connect.hku.hk lixiang960927@gmail.com
}

\begin{document}
\maketitle
\input{sec/0_abstract}

\input{sec/1_intro}

\input{sec/2_related_work}
\input{sec/3_evaluation_metric}

\input{sec/4_dataset}

\input{sec/5_experiment}
\input{sec/6_conclusion}

\section*{Acknowledgements}

This work was supported by the National Natural Science Foundation of China (62476011).

\input{sec/X_suppl}
\clearpage
{
    \small
    \bibliographystyle{ieeenat_fullname}
    \bibliography{main.bib}
}

\end{document}

%% file: sec/0_abstract.tex
\begin{abstract}

\noindent Recently, significant advances have been made in Video Large Language Models (Video LLMs) in both academia and industry. However, methods to evaluate and benchmark the performance of different Video LLMs, especially their fine-grained, temporal visual capabilities, remain very limited. On one hand, current benchmarks use relatively simple videos (e.g., subtitled movie clips) where the model can understand the entire video by processing just a few frames. On the other hand, their datasets lack diversity in task format, comprising only QA or multi-choice QA, which overlooks the models' capacity for generating in-depth and precise texts. Sports videos, which feature intricate visual information, sequential events, and emotionally charged commentary, present a critical challenge for Video LLMs, making sports commentary an ideal benchmarking task. Inspired by these challenges, we propose a novel task: sports video commentary generation, developed \textbf{SCBench} for Video LLMs. To construct such a benchmark, we introduce (1) \textbf{SCORES}, a six-dimensional metric specifically designed for our task, upon which we propose a GPT-based evaluation method, and (2) \textbf{CommentarySet}, a dataset consisting of 5,775 annotated video clips and ground-truth labels tailored to our metric. Based on SCBench, we conduct comprehensive evaluations on multiple Video LLMs (e.g. VILA, Video-LLaVA, etc.) and chain-of-thought baseline methods. Our results found that InternVL-Chat-2 achieves the best performance with 5.44, surpassing the second-best by 1.04. Our work provides a fresh perspective for future research, aiming to enhance models' overall capabilities in complex visual understanding tasks. Our dataset will be released soon.

\end{abstract}

\label{sec:abstract}

%% file: sec/1_intro.tex
\section{Introduction}
\label{sec:intro}

\begin{figure} 
    \centering 
    \begin{minipage}{0.98\textwidth} 
        \centering 
        \includegraphics[width=\textwidth]{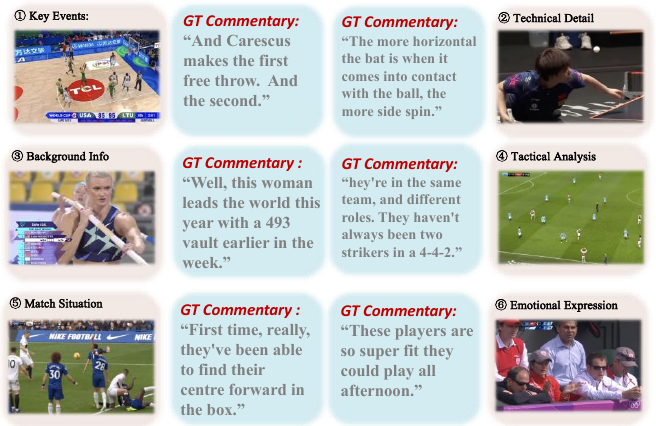} 
    \end{minipage}
    \vspace{1em} 
    \vspace{-2mm}
    \caption{\textbf{A sample from Six-Dimensional Metric and CommentarySet}, including six dimensions' description, frames from videos, and corresponding ground truth commentary.} 
    \label{Fig.intro_visualize} 
    \vspace{-5mm}
\end{figure}

With the advancement of Video Large Language Models (Video-LLMs), neural networks are capable of processing video sequences to perform multi-modality tasks based on temporal and spatial information. Recent advancements have facilitated significant improvements in the automatic generation of video content summaries, captions, and even interactive commentary~\cite{miech2019howto100m,zhou2024survey,lin2024vilapretrainingvisuallanguage,liu2024llavanext}. 

 %MLLM benchmark limitation
 \noindent To holistically assess their margin of improvement, many benchmarks and evaluation methods have been proposed~\cite{ning2023video,sanders2024survey,ning2023videobenchcomprehensivebenchmarktoolkit,fu2024videommefirstevercomprehensiveevaluation}. Existing benchmarks for Video-LLMs predominantly concentrate on general video understanding tasks\cite{ning2023video,sanders2024survey}, such as action recognition~\cite{mangalam2023egoschemadiagnosticbenchmarklongform} and long context video understanding~\cite{wang2024lvbenchextremelongvideo,fang2024mmbenchvideolongformmultishotbenchmark}.
 However, these methods mostly capture image-level understanding tasks that mostly test the model's abilities in the spatial understanding tasks in the form of QA. Consequently,~\cite{li2024mvbenchcomprehensivemultimodalvideo} have curated tasks that require reasoning spanning over the entire videos for MLLMs. However, this benchmark still suffers from the lack of dense events, where Gemini extracts frames at 1 FPS is sufficient to perform well in their experiments, meaning that the density of events is not enough where most videos can perform well using only long intervals of video frames. Additionally, there is a lack of diversity in task format, comprising only QA or multi-choice QA, which overlooks the models' capacity for generating in-depth and precise texts, thus precluding a deeper evaluation of the models' analytical and interpretative abilities. 

%Sports Video Value
\noindent Discovering limitations in existing datasets, we turn to sports videos, especially at the professional level, to present a challenging domain for video understanding for Video-LLMs. These videos involve complex human-object interactions, strategic gameplay, and visually demanding tasks like event and motion spotting. The rapid shifts between high-action sequences and static moments require models to prioritize information effectively, while the need to interpret intricate motions and maintain context over time adds to the complexity, especially for real-time sports commentary.

 \noindent Although many existing sports video datasets have been explored, they are not well-suited for evaluating video-based multimodal large language models (MLLMs), mainly due to their lack of sports diversity and narrow aspect of the video. The current domain of sports video analysis has been a rich and multifaceted domain that covers several visual tasks, ranging from re-id~\cite{comandur2022sportsreidimprovingreidentification}, OCR~\cite{solberg2024playertvadvancedplayertracking}, temporal localization~\cite{liu2022fineaction}, fine-grained action classification~\cite{liu2024crossblockfinegrainedsemanticcascade}, motion understanding~\cite{feng2024chatpose}, etc. 
  These tasks often focus on narrow tasks or evaluate only specific aspects of the content, without considering the full spectrum of multimodal information. For instance, in tasks such as sports video captioning, the focus is typically limited to key event captions, such as describing specific actions performed by athletes, while neglecting other critical aspects including emotional context and background information. 
We have identified sports video commentary as a highly suitable and efficient task for evaluating the capabilities of video MLLMs. Sports commentary generation, unlike previous approaches, provides comprehensive coverage of sports video information by capturing key events, tactical analysis, emotional expressions, and background information. Moreover, the abundance of existing sports commentaries makes it easier to scale up without the need for extensive manual annotation.

\noindent To this end, we propose the fine-grained CommentarySet, a novel and comprehensive dataset entailing a new commentary task designed to better utilize the diverse information embedded in sports videos. As demonstrated by Fig.~\ref{Fig.intro_visualize}, CommentarySet consists of 5,775 high-quality sports video clips spanning six diverse sports including athletics, basketball, soccer, gym, table tennis, and tennis, each meticulously annotated with professional English commentary. This dataset serves as a robust benchmark for evaluating models in the domain of sports commentary, offering a rich variety of event types, commentary styles, and detailed labels. Furthermore, we notice that the task of sports commentary is significantly different from traditional video captioning. Thus, we introduce a novel six-dimensional commentary with human annotation. Our dataset and evaluation methods account for the dynamic and emotionally resonant nature of sports commentary, which requires not only accurate event descriptions but also a deep integration of contextual information. Based on such inputs, we introduce a baseline method with Chain-of-Thought to perform video understanding and narration in two steps and evaluate its limitations. Additionally, we present a GPT-based evaluation method grounded in our six-dimensional metric, SCORES, as shown in Fig.~\ref{Fig.intro_visualize}, offering a comprehensive and nuanced framework for evaluating sports video commentary.
In our extensive experiments, we find all multi-modal video models failing to perform successful sports commentary, remaining a notably challenging domain.

\noindent Our main contributions are summarized as follows:

\begin{itemize}
    \item  We introduce CommentarySet, the first dataset featuring professionally annotated sports video commentary across multiple disciplines, encompassing six sports and over ten international competitions.
    
    \item The dataset includes meticulously designed six-dimensional labels, utilizing the diverse information of sports commentary to enable robust evaluation of advanced video language models. These account for the dynamic and emotionally resonant nature of sports commentary, which requires not only high-frequency accurate event descriptions but also a deep integration of contextual information.

    \item To validate existing approaches on our dataset, we developed a baseline utilizing chain-of-thought to first understand the video and classify the type of response, and then generate the commentary. However, we find such approaches still distant from practical commentary.
    
    \item We evaluate several open-source pre-trained Video LLMs as baselines and fine-tune Video-LLaVA and Chat-UniVi-1.5 on CommentarySet, boosting their performance by ~8\% and ~6\%, respectively. SCORES demonstrates significant limitations in existing models' capabilities in performing sports commentary generation.
  
\end{itemize}

%% file: sec/2_related_work.tex
\section{Related Works}
\label{sec:related_work}
\subsection{Benchmarking Video LLMs}
Recently, the emergence of numerous Video LLMs, such as VILA-1.5~\cite{lin2024vilapretrainingvisuallanguage}, LLaVA-NeXT-Video~\cite{liu2024llavanext}, Video-LLaVA~\cite{lin2023videollavalearningunitedvisual}, InternVL~\cite{chen2024internvlscalingvisionfoundation} and LongVA~\cite{zhang2024longcontexttransferlanguage}, have been proven to be effective in solving various downstream tasks. Consequently, various benchmarks, based on different evaluation dimensions, have been introduced to evaluate their performance. Previous works like Perception Test~\cite{pătrăucean2023perceptiontestdiagnosticbenchmark}, EgoSchema~\cite{mangalam2023egoschemadiagnosticbenchmarklongform}, and TempCompass~\cite{liu2024tempcompassvideollmsreally} have focused on specific aspects of the models' capabilities, such as spatial understanding and temporal comprehension. More recent works, including Video-Bench~\cite{ning2023videobenchcomprehensivebenchmarktoolkit}, Video-MME~\cite{fu2024videommefirstevercomprehensiveevaluation}, MMBench-Video~\cite{fang2024mmbenchvideolongformmultishotbenchmark}, MVBench~\cite{li2024mvbenchcomprehensivemultimodalvideo} and LVBench~\cite{wang2024lvbenchextremelongvideo}, evaluate models by defining task-specific hierarchical capability taxonomy of the model and adopts longer videos. However, current works do not focus on evaluating the ability to understand more comprehensive and less perceptible visual information and fail to evaluate their temporal memory ability. Besides, their metrics are based on the explicit features or QA pairs, which fall short of the ability to evaluate our task with implicit conceptions.

\subsubsection{Sports Video Datasets}
Sports videos for video captioning models represent a branch within video understanding tasks, with numerous datasets and benchmarks proposed for various sports and tasks. Datasets such as TenniSet~\cite{Faulkner2017TenniSetAD}, FineGym~\cite{shao2020finegymhierarchicalvideodataset}, P2A~\cite{bian2024p2anetdatasetbenchmarkdense}, and SoccerNet~\cite{deliège2021soccernetv2datasetbenchmarksholistic} provide detailed labels for technical actions and events in the games. Specifically, datasets like TenniSet, FineGym, and P2A contain a large number of standardized labels about on-field events or the description of on-field player actions. Another type of research focuses on interactions between athletes, exemplified by SportsHHI~\cite{wu2024sportshhidatasethumanhumaninteraction}, which focuses on athletes' location and interaction classification tasks for sports videos. Although these datasets provide data for sports video recognition tasks of Video LLMs, there is still a lack of commentary-based sports videos, which is exactly what our newly proposed dataset addresses.\\
\subsubsection{Video Captioning on Sport Videos}
In the field of sports video, the datasets mentioned in the previous section also serve as datasets for the sports video captioning task, offering real-time labels for sports videos. The datasets for these captioning tasks provide time stamps and specific captions for the video clips which Video LLMs are required to generate captions for. The structure of a dataset directly influences the evaluation dimensions and the methods used. Sports datasets like SoccerNet~\cite{deliège2021soccernetv2datasetbenchmarksholistic} and SportsHHI~\cite{wu2024sportshhidatasethumanhumaninteraction} have evaluated models' abilities in event understanding or interaction recognition tasks within sports videos. These captioning tasks assess models from a textual perspective such as feature or N-gram similarity. Despite advancements in datasets and benchmarks for technical actions and interactions, there remains a need for a comprehensive sports dataset and benchmark for commentary-based captioning - whose particularity has been mentioned above. That's why we've come up with a new commentary benchmark to measure this ability.

%% file: sec/3_evaluation_metric.tex
\section{Evaluation \& Metric}
\label{eval&metric}

In this section, we first define the task of sports commentary, highlighting its distinctions from conventional video captioning tasks. Subsequently, we present our novel six-dimensional metric, SCORES, tailored for this task, compared with traditional metrics and the vanilla GPT-based method. Finally, we provide an in-depth description of our GPT-based evaluation method, grounded in the six-dimensional metric.

\subsection{Task Definition}

\noindent Sports commentary provides descriptions, analysis, and insights to enhance the viewing experience with detailed information, tactical breakdowns, and highlights. The task involves providing a video input and a prompt to the model. The demanding output is the corresponding commentary.

\begin{figure}[htbp] 
    \centering
    \begin{subfigure}[b]{0.8\textwidth} 
        \centering
        \includegraphics[width=\textwidth]{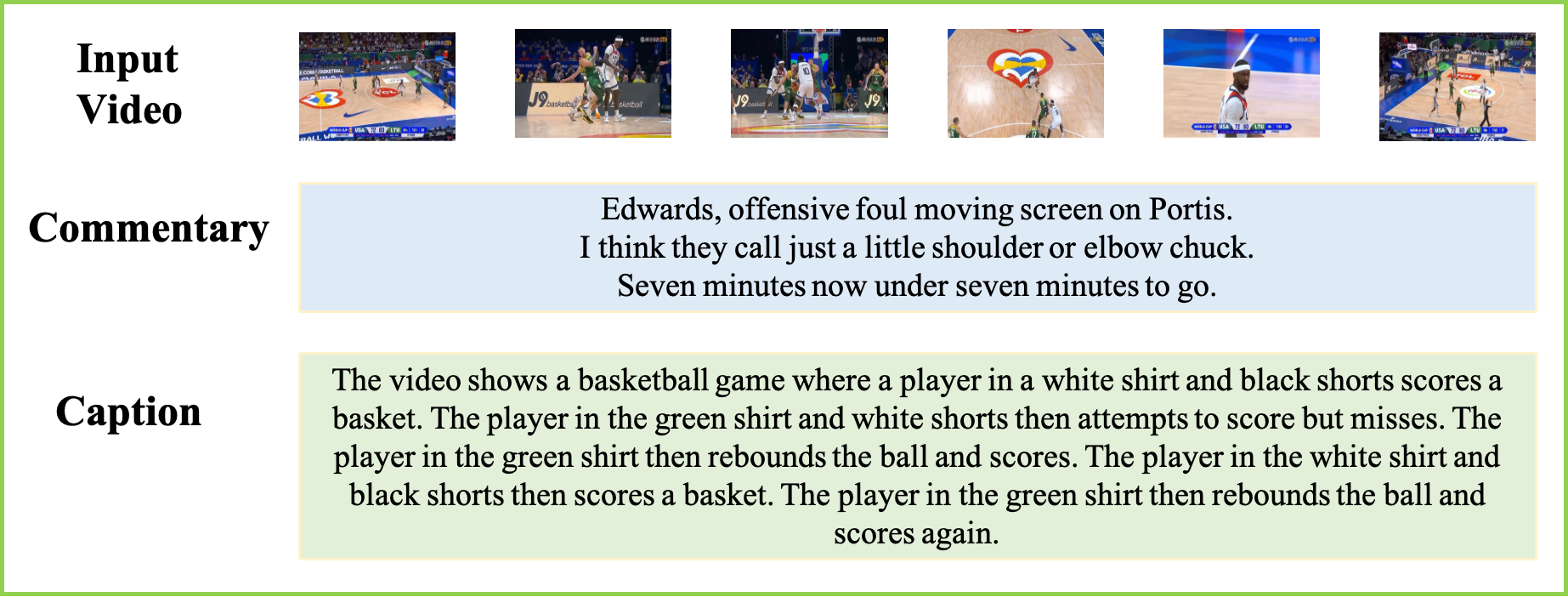} 
    \end{subfigure}
    \vspace{5mm} 
    \begin{subfigure}[b]{0.8\textwidth} 
        \centering
        \includegraphics[width=\textwidth]{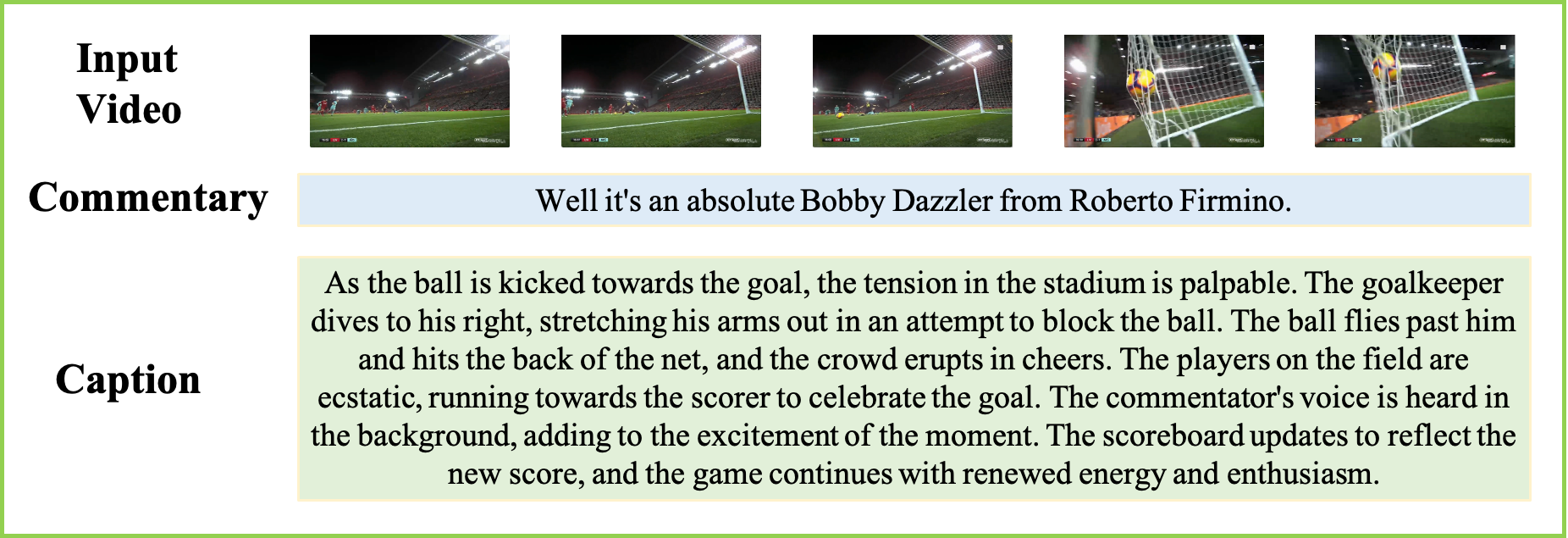} 
    \end{subfigure}
    \caption{\textbf{Distinctions between Caption and Commentary.} Although captioning provides precise visual descriptions, commentary delivers dynamic, context-rich, and emotionally resonant narration.}
    \label{Fig.twocap}
    \vspace{-5mm}
\end{figure}

\noindent Unlike traditional video captioning, which only focuses on describing visual content, sports commentary requires not only accurate event descriptions but also timely, emotionally resonant narration that aligns with the game's pace and context. For example, when game intensity is relatively low, commentators may provide tactical analysis or details of the players or the teams, while remarkable plays might trigger emotionally charged reactions like "What a goal!" and "Amazing!" Fig.~\ref{Fig.twocap} illustrates the difference between caption and commentary.

\subsection{SCORES}

\noindent Having established the definition of our task and distinguished it from video captioning, we now explain the rationale for our novel metric. We first compare it with traditional captioning metrics to demonstrate its necessity, followed by a comprehensive introduction to the six-dimensional metric.

\subsubsection{Why Propose a New Metric}

\noindent Existing benchmarks for Video LLMs mostly deploy metrics designed for either QA or multi-QA tasks, which are unsuitable for our task. Meanwhile, the metrics currently used for video captioning tasks fail to fully accommodate our specific requirements.

\noindent To be more specific, traditional metrics (e.g. \textbf{BLEU}~\cite{papineni-etal-2002-bleu}, \textbf{ROUGE\_L}~\cite{lin-2004-rouge}, \textbf{METEOR}~\cite{METEOR}, \textbf{CIDEr}~\cite{cider} and \textbf{SPICE}~\cite{SPICE}) primarily evaluate the lexical or structural similarity between generated texts and reference texts, focusing on aspects like n-gram overlap, sequence matching, and semantic accuracy. However, they do not capture the model's ability to understand fine-grained professional details, temporal dynamics, or human emotions in commentary. Apart from that, compared to the vanilla GPT approach, which directly evaluates the generated commentary based on ground truth, our method not only provides a more detailed and fine-grained scoring standard for GPT but also allows for a more intuitive demonstration of the model's capabilities across different dimensions.\\

\noindent Detailed experimental demonstrations of comparison are included in the Sec.~\ref{subsubsec:quanti_ana}.

\subsubsection{Details of SCORES}

\noindent To address the limitations of traditional metrics and better evaluate the multifaceted nature of sports commentary, we propose our novel six-dimensional metric, \textbf{SCORES} (\textbf{S}ituation, ta\textbf{C}tic, em\textbf{O}tion, backg\textbf{R}ound, key \textbf{E}vents, technique\textbf{S}), to be specific:\\
1. \textbf{Key Events Caption:} This dimension assesses the model’s ability to accurately detect and describe significant events, such as scores, turnovers, fouls, and substitutions, reflecting its proficiency in real-time event detection and textual articulation.\\
2. \textbf{Technical Detail Analysis:} It evaluates models’ capacity to analyze and explain player actions, such as passing and shooting techniques, reflecting its fine-grained visual understanding capability and contextual knowledge integration.\\
3. \textbf{Background Information Interpretation:} This dimension measures the model's ability to integrate visual and textual information about players' and teams' histories, characteristics, and performance, which demonstrates its contextual knowledge integration and deep understanding of visual information(e.g. game intensity and event frequency).\\
4. \textbf{Tactical Analysis:}  It assesses the model's skill in interpreting and explaining team strategies, formations, and in-game tactical adjustments by analyzing visual inputs and conveying these insights through commentary, highlighting its temporal understanding of game dynamics.\\
5. \textbf{Match Situation Interpretation:} This dimension evaluates the model's ability to interpret the current state of the match, such as the score and momentum, through visual cues and predict its likely progression, providing dynamic situational insights.\\
6. \textbf{Emotional Expression:} It measures the model’s ability to detect emotional elements from visual data, such as player reactions and crowd energy, and effectively express these emotions in text to engage the audience emotionally.

\begin{figure*} 
    \centering 
    \begin{minipage}{0.90\textwidth} 
        \centering 
        \includegraphics[width=\textwidth]{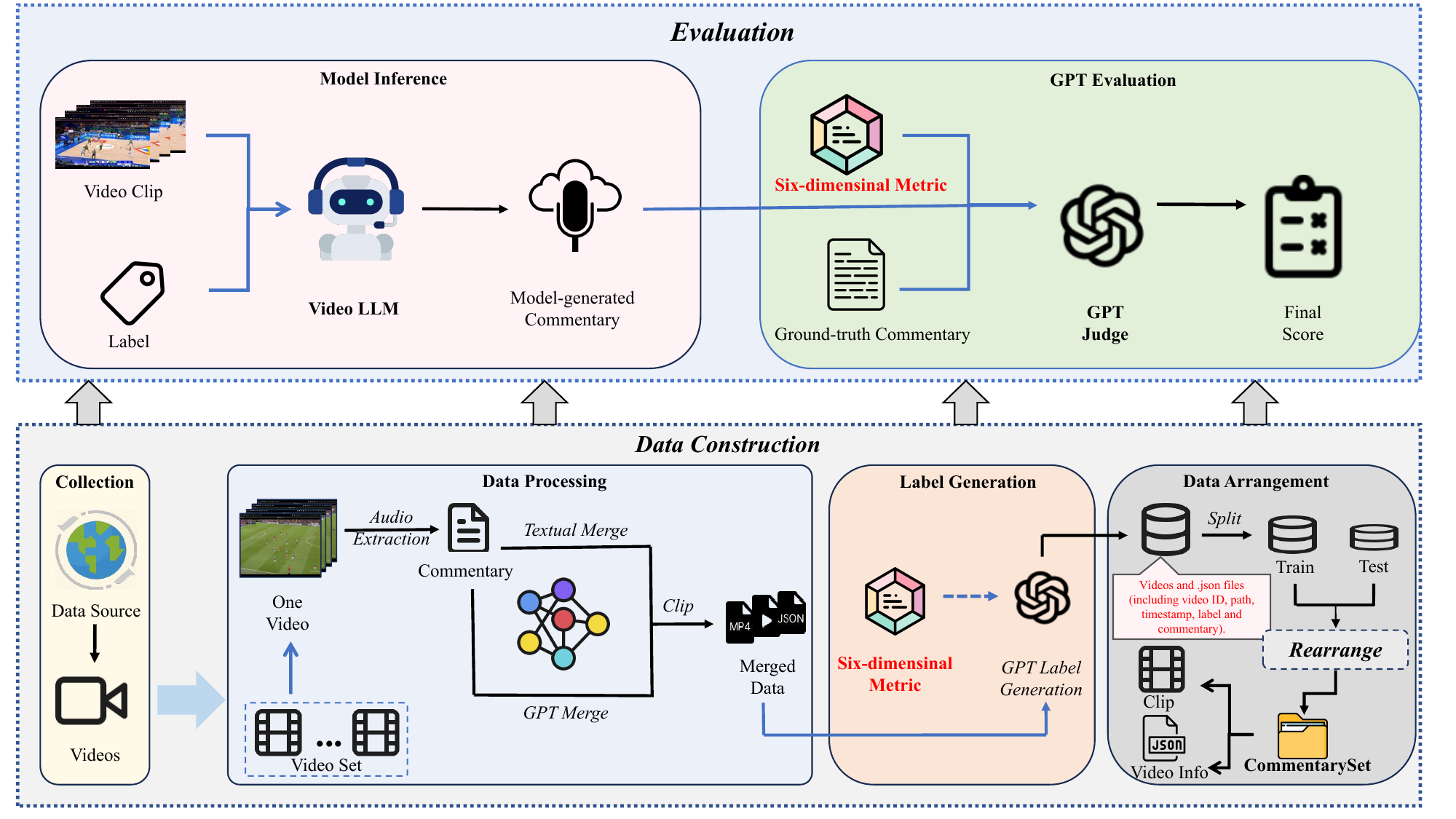} 
    \end{minipage}
    \vspace{-2mm}
    \caption{\textbf{The overall framework of SCBench}. SCBench consists of two parts: the construction of CommentarySet and our GPT-based evaluation.} 
    \label{Fig.pipeline} 
    \vspace{-2mm}
\end{figure*}

\subsection{GPT-based Evaluation}
\label{sec:eval}

\noindent Built upon the SCORES, we propose a novel GPT-based evaluation method tailored for sports video commentary, as shown in Fig.~\ref{Fig.pipeline}. Before evaluation, we need to generate labels for our data according to the six-dimensional metric, which is introduced in Sec.~\ref{data_construct}. Here, we provide a detailed overview of evaluation processes.

\begin{table*}
    \centering
    \caption{Commentary statistics of our proposed dataset across different sports domains.}
    \resizebox{1\textwidth}{!}{\begin{tabular}{l|ccccc}
    \toprule
        Domain &  Clip Num &  Avg Duration &  Commentary Length &  Commentary Freq & Source \\
    \midrule
        Total &  5,775 &  14.5s &  30.14 &  45.73 & -\\
        Athletics &  1,383 &  18.38s &  44.77 &  30.44 &  IAAF Gold Label Road Races, IAAF Diamond League\\
        Basketball &  1,613 &  12.06s &  23.36 &  26.18 & 2023 FIBA Basketball World Cup\\
        Soccer &  696 &  10.26s &  19.79 &  70.51 & English Premier League\\
        Gym &  1,153 &  16.49s &  34.68 &  67.61 & 2015/2016 World Gymnastics Championship, etc. \\
        Table Tennis &  719 &  9.96s &  20.63 &  31.7 & 2019-2022 World Cups\\
        Tennis &  211 &  26.27s &  27.86 &  141.92 & 2012 London Olympics\\
    \bottomrule
    \end{tabular}}
    \label{tab:info}

\end{table*}

\begin{figure}[htbp]
    \vspace{-2mm}
    \centering
    \begin{minipage}{0.46\textwidth} 
        \centering
        \captionof{table}{Comparison of different datasets across different sports domains. \textbf{Avg Duration} is the average duration of all clips in the sport}
        \resizebox{\textwidth}{!}{\begin{tabular}{l|cccc}
        \toprule
            Dataset &  Domain &  Clip Numbers &  Avg Duration\\
        \midrule
            TennisSet &  Tennis &  3,568 &  1.08s  \\ %~\cite{faulkner2017tenniset}
            FineGym &  Gymnastic &  4,885 &  8s \\ %~\cite{shao2020finegym}
            FSN &  Basketball &  2,000 &  - \\ %~\cite{yu2018fine}
            SoccerNet-caption &  Soccer &  942 &  238s \\ %~\cite{mkhallati2023soccernet}
            CommentarySet (ours) &  Multiple sports &  5,775 &  14.5s  \\
        \bottomrule
        \end{tabular}}
        \label{tab:compare}
    \end{minipage}
    \hfill
    \begin{minipage}{0.48\textwidth} 
        \centering
        \includegraphics[width=\textwidth]{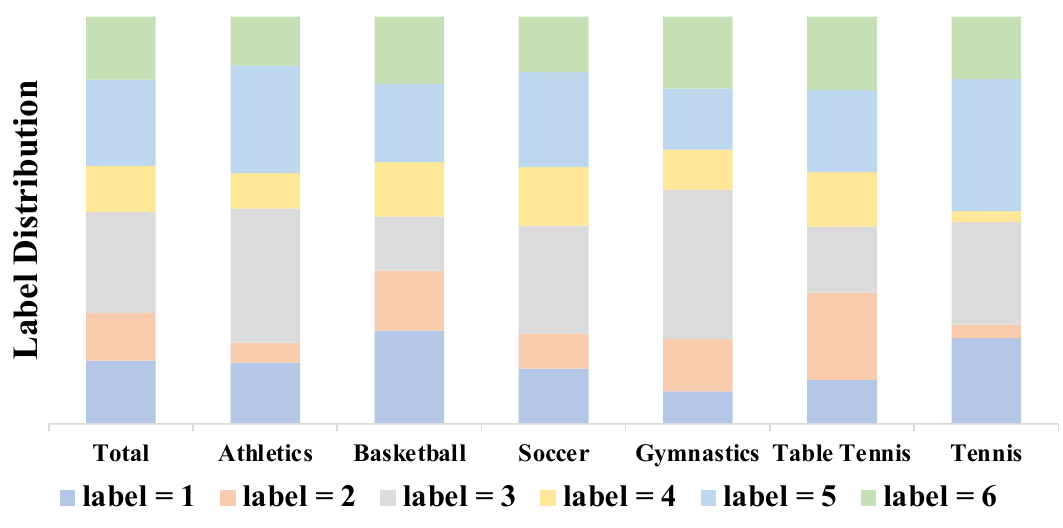}
        \caption{Distribution of the labels}
        \label{Fig.label_distribution}
    \end{minipage}
    \vspace{-2mm}
\end{figure}

\noindent The evaluation process could be divided into two stages: Model Inference and GPT Evaluation. In the first stage, we employ a three-tier prompt structure of \textbf{ task definition - corresponding metric definition - task requirement}: For each video clip, we first provide the model with the definition of tour task, followed by the definition of the SCORES label for that clip. Finally, we input the task requirement into the model. As for \textbf{GPT Evaluation}, we input both the ground truth and model-generated commentary into our GPT judge, instructing it to evaluate the generated commentary based on the ground truth and the perspectives of the labels. The evaluation yields a score from 0 to 10, with 0 indicating very poor quality and 10 representing perfection.

%% file: sec/4_dataset.tex
\section{Dataset}
\label{sec:dataset}
Our goal is to advance research in the field of sports video commentary by introducing a challenging benchmark with professional commentary annotations. To this end, we have developed the fine-grained \textbf{CommentarySet}, which contains 5775 high-quality(1080P) sports video clips. Each video is well annotated, and the overall dataset structure will be discussed in the next section.\\

\subsection{Dataset Construction}
\label{data_construct}
\noindent \textbf{Data Structure}
To better evaluate sports video commentary, we create a novel dataset called CommentarySet for SCBench, containing a total of 5,775 clips, with 4,908 in the training set and 867 in the test set, the comparison with other datasets is shown in Tab.~\ref{tab:compare}. Part of the clips is selected from FineGym~\cite{shao2020finegymhierarchicalvideodataset} and TenniSet~\cite{Faulkner2017TenniSetAD}. Specifically, CommentarySet includes clips from six types of sports, along with their corresponding timestamps in the original videos. The necessary information on the six sports we selected is shown in Tab.~\ref{tab:info}.  Moreover, each clip is accompanied by carefully selected English commentary text, which can serve as a ground truth label in sports commentary evaluation. To implement the evaluation metrics proposed in this paper, we also provide each clip with precise six-dimensional labels. Additional information about CommentarySet is shown in supplementary material.\\

\noindent \textbf{Dataset Creation Process}
To create CommentarySet and ensure the reliability, professionalism, and accuracy of the video resources and commentary, we follow the pipeline below for dataset creation, as shown in Fig.~\ref{Fig.pipeline}:\\
\textbf{1. Collection:} We collect a large number of high-definition sports videos with original English commentary from various sports and events available online and extracted the English commentary from the original audio, preserving their corresponding timestamps.\\
\textbf{2. Merging:} The commentary sentences obtained from the Collection process are fragmented. So, we add a process to merge them. To be specific, let the previous and current commentary be Com\(_1\) and Com\(_2\) respectively, and let their time stamps be [b\(_1\), e\(_1\)] and [b\(_2\), e\(_2\)]. The merged commentary and its time stamp can be obtained using the following formula: \\
\begin{equation}
    \text{New Com} = \text{Com}_{1} + \text{Com}_{2}
\end{equation}
\begin{equation}
    \text{New Time Stamp} = [b_{1}, e_{2}]
\end{equation}
Regarding the determination of whether two commentaries should be merged, we conduct the following two-stage decision process:\\
Stage 1 Textual Merging: We use the sentence encoder to encode the sentences and then apply cosine similarity to merge commentary with high textual similarity such as repetitive cheers or consecutive sentences describing the same content. To be specific, the similarity (Sim) is calculated using the cosine similarity. After obtaining Sim, we simply decide whether to merge according to the following formula:
    \begin{equation}
        Action =
        \begin{cases}
        \text{Merge} & \text{if Sim } > \text{ threshold} \\
        \text{Separate} & \text{else}
        \end{cases}
    \end{equation}
    where the threshold is set to 0.7 as our experimental choice.\\
Stage 2 GPT Merging: Some segments with continuous semantic meaning but no textual overlap remained. We use GPT-4o-mini to judge whether to merge consecutive commentary segments, the prompt would be shown in supplementary material. This is applied to segments where there exists a logical connection and describes the same topic.

\noindent  \textbf{3. Slicing:} Before slicing, we manually refine the timestamps. The full sports match videos are sliced according to the timestamps, generating clip-commentary pairs.\\
\noindent  \textbf{4. Six-Dimensional Label Generation:} Finally, we use GPT-4o-min to classify the semantic content of the commentary. Each commentary sentence is categorized into one or more of the six dimensions mentioned in the metric section.

\subsection{Statistic Analysis \& Comparison}
This section will primarily present specific statistical information about CommentarySet. We will also demonstrate that CommentarySet is a more suitable dataset for our proposed SCBench compared to other existing sports video captioning datasets, offering greater comprehensiveness and accuracy. Tab.~\ref{tab:info} fully showcases the specific parameters within CommentarySet.\\

\noindent \textbf{Diversity} The six sports included in CommentarySet are carefully selected to ensure strong diversity. As shown in Tab.~\ref{tab:info}, the six selected sports exhibit significant diversity in terms of match speed, player density, commentary length, and commentary frequency. For instance, the average duration of table tennis clips is the shortest at just \textbf{9.96s}, while athletics, gymnastics, and tennis have clip lengths \textbf{1.5} to \textbf{2} times longer, reflecting faster match paces and shorter commentary content. In terms of commentary frequency, basketball has the highest frequency, while soccer and tennis have relatively lower frequencies, correlating closely with the frequency of events occurring on the field.\\

\noindent \textbf{Commentary Style} Commentary style is another critical aspect. The content and focus of commentary vary across different sports, which motivates our proposal of SCORES. Through the analysis of the distribution of the six labels across sports, as shown in Fig.~\ref{Fig.label_distribution}, we have observed distinct commentary styles. For example: In basketball, key event captions account for 44.33\%, indicating continuous on-field events; In gymnastics, 61.84\% of the commentary includes Background Information Interpretation, as commentators frequently introduce each participating athlete. The diversity of sports leads to diverse commentary styles, which is quantitatively reflected in the distribution of labels across different sports. CommentarySet includes data on commentary styles across different sports, serving as a foundational reference for commentary generation tasks, a unique statistical concept compared to other datasets.

%% file: sec/5_experiment.tex
\begin{table*}
    \centering
    \caption{\textbf{Performance of Video LLMs.} Including the pre-trained models and the fine-tuned models. "ICL" indicates utilizing the ICL fine-tuning dataset while "CoT" indicates using the CoT training set.}
    \resizebox{\textwidth}{!}{\begin{tabular}{l|c|cc|cccccc|c}
    \toprule
    \multirow{2}{*}{\textbf{Model Name}} & \multirow{2}{*}{\begin{tabular}[c]{@{}c@{}}\textbf{Model}\\\textbf{Params}~~\end{tabular}}
    &\multicolumn{2}{c}{\textbf{Traditional Metric}}& \multicolumn{6}{c}{\textbf{SCORES in each sport event (0-10)}} & \multirow{2}{*}{\textbf{SCORES (0-10)}} \\
    \cmidrule{3-10}
     & & BLEU(0-100) & CIDEr(0-1) & Table Tennis & Basketball & Soccer & Gym & Tennis & Athletics\\
    \midrule
    Mini-InternVL-Chat-4B-v1.5 & 4B & 0.44 & \textbf{0.12} & 4.98 & 4.42 & 4.18 & 4.36 & 4.54 & 4.20 & 4.40 \\
    Video-LLaVA & 7B & 0.67 & 0.08&  2.65 &  2.56 &  2.33 &  2.97 &  2.17 &  2.25 &  2.55 \\
    LongVA & 7B & 0.39 & 0.1 & 4.80 &  3.59 &  3.59 &  3.79 &  3.83 &  3.82 &  3.85 \\
    LLaVA-NeXT-Video & 7B & 0.35 & 0.1 & 4.77 &  4.39 &  4.22 &  4.24 &  4.28 &  3.65 &  4.20 \\
    Chat-UniVi-1.5 & 7B & 0.56 & 0.11 & 3.67 &  3.48 &  2.96 &  3.68 &  2.70 &  3.08 &  3.37 \\
    Kangaroo & 8B & 0.83 & 0.01 & 1.60 &  1.64 &  1.70 &  1.97 &  2.02 &  1.89 &  1.79 \\ 
    VILA & 34B & 0.66 & 0.09 & 3.83 &  3.42 &  3.24 &  3.78 &  3.00 &  2.92 &  3.40 \\ 
    InternVL-Chat-2 & \textbf{40.1B} & 0.32 & 0.11 & \textbf{5.92} & \textbf{5.50} & \textbf{5.42} & \textbf{5.31} & \textbf{4.98} & \textbf{5.29} & \textbf{5.44} \\ 
    \midrule
    Video-LLaVA-ICL & 7B & 2.11 & 0.06 & 2.76 & 2.70 & 2.16 & 3.01 & 2.98 & 2.60 & 2.70 \\
    Chat-UniVi-1.5-ICL & 7B & \textbf{2.57} & 0.06 & 3.66 &  3.57 &  3.03 &  3.79 &  2.81 &  3.04 &  3.42 \\
    Video-LLaVA-CoT & 7B & 1.51 & 0.05 & 2.40 & 3,00 & 2.61 & 2.77 & 3.64 & 2.67 & 2.77 \\
    Chat-UniVi-CoT & 7B & 1.44 & 0.06 & 2.44 & 3.78 & 3.47 & 3.92 & 3.42 & 3.62 & 3.57 \\
    \bottomrule
    \end{tabular}}
    \label{exp}
\end{table*}

\section{Experiments}

\label{sec:exp}

We conduct extended experiments which could be divided into two main parts. First, we evaluated the zero-shot commentary capabilities of multiple open-source Video LLMs on our SCBench. Subsequently, we assessed the effects of ICL~\cite{brown2020languagemodelsfewshotlearners} and CoT~\cite{wei2023chainofthoughtpromptingelicitsreasoning} methods on fine-tuning Video-LLaVA and Chat-UniVi, which are the baseline models. In this section, we first introduce our experiment settings, including model selection and hyperparameters. Then we present quantitative results for both zero-shot experiments and fine-tuned models. Finally, we present a detailed visualization, analysis, and case study of the experimental results across different models and fine-tuning methods.

\subsection{Settings and Configurations}

\noindent All the experiments are conducted on 8 A800 GPUs. We perform evaluation on eight open-source pre-trained Video LLMs, including VILA~\cite{lin2024vilapretrainingvisuallanguage}, Video-LLaVA~\cite{lin2023videollavalearningunitedvisual}, LongVA~\cite{zhang2024longcontexttransferlanguage}, LLaVA-NeXT-Video~\cite{liu2024llavanext}, Kangaroo~\cite{liu2024kangaroopowerfulvideolanguagemodel}, Chat-UniVi-1.5~\cite{jin2024chatuniviunifiedvisualrepresentation}, InternVL-2~\cite{InternVL2}, and Mini-InternVL-v1.5~\cite{gao2024miniinternvlflexibletransferpocketmultimodal}, as well as two fine-tuned models, which are Video-LLaVA and Chat-UniVi-1.5.\\

\noindent As is illustrated in the section~\ref{sec:eval}, the evaluation process is divided into two main stages: Model Inference and GPT Evaluation. In the Model Inference stage, we follow their official configurations and try to use more frames and more maximum output tokens for commentary generation. To be more specific, for InternVL-2, Video-LLaVA, VILA, and Kangaroo, the maximum number of new tokens is set to 512. For the number of sampled frames, both LLaVA-NeXT and LongVA are set to 32 frames. For GPT Evaluation, we utilize \textit{GPT4o-mini} model as our judge, with the temperature set to 0.1. As for fine-tuning, we follow the official document for full-parameter tuning and set the learning rate marginally lower. Detailed settings and prompts are shown in the supplementary material. 

\begin{figure} 
    \centering 
    \vspace{-2.5mm}
    \begin{minipage}{0.8\textwidth} 
        \centering 
        \includegraphics[width=\textwidth]{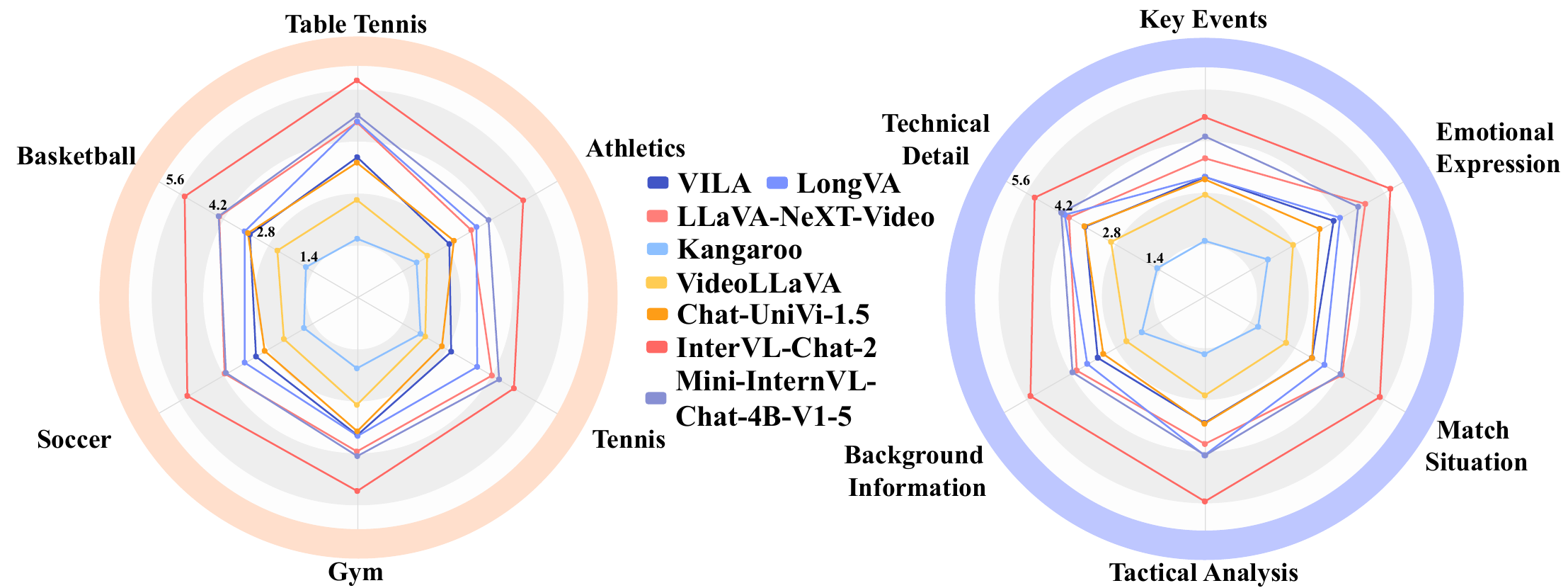} 
    \end{minipage}
    \vspace{-0.5mm}
    \caption{Model Performances on Six Sports (left) and Six-Dimensions in SCORES (right) tasks} 
    \label{Fig.radar} 
    \vspace{-5mm}
\end{figure}

\subsection{Quantitative Results}

\textbf{Zero-shot Model Performance} To comprehensively evaluate the performance on our SCBench, we conduct extensive experiments on the eight aforementioned pretrained Video LLMs, the result is shown in Tab.~\ref{exp}. Overall, InternVL-2, with \textbf{40.1B} parameters, perform the best, achieving a score of \textbf{5.44}, surpassing second-best, Mini-InternVL-Chat-v1.5, by \textbf{1.04}. Following this, LLaVA-NeXT-Video, LongVA and VILA score \textbf{4.20}, \textbf{3.85} and \textbf{3.40}, respectively. Our evaluation is based on SCORES. So we calculate the models' performance in all six dimensions and create a six-dimensional radar chart, as shown in Fig.~\ref{Fig.radar}, showing the comprehensive performance of models. Additionally, as the models show varying strengths across different sports, we also create a six sports radar chart in Fig.~\ref{Fig.radar} to show models' performance across different sports types. For example, Video-LLAVA excels in gymnastics, outperforming its score in table tennis by \textbf{0.32} and LongVA stands out in the table tennis task, achieving a score of \textbf{4.80}, but falls short in other types of sports. The strengths across different sports highlight the models' specific capabilities mentioned in the dataset section.\\
\noindent \textbf{ICL \& CoT Fine-tuning} We propose two methods of fine-tuning and evaluate them with Video-LLaVA and Chat-UniVi on our SCBench:

\noindent 1. \textbf{In-Context Learning (ICL) Fine-tuning}: In-Context Learning refers to the ability to perform specific tasks by leveraging contextual information embedded within the prompt. In our fine-tuning, We structure the fine-tuning QA setup similar to the prompt format used in Model Inference, instructing the model to directly generate commentary based on the video, task definition, and the perspectives of the ground truth metrics, intended to enhance the model’s comprehension of the task and the metric perspectives and align the characteristics of the metric dimensions with the generated commentary.

\noindent 2. \textbf{Chain-of-Thought (CoT) Fine-tuning}: Chain-of-Thought (CoT) is a prompting technique that enhances models' reasoning by breaking down complex problems into smaller steps. In fine-tuning, we format the training data to require the model to analyze the video, task definition, and all metric perspectives to first identify the relevant perspective before generating commentary. This approach, which is more challenging than providing a specific metric dimension, aims to encourage the model to emulate human reasoning in commentary, enhancing coherence and conciseness.

\noindent The scores of fine-tuned models are demonstrated in Tab.~\ref{exp}. The results indicate that both fine-tuning methods enhance the model's performance, with CoT providing a more substantial improvement. Specifically, CoT increased performance by \textbf{8\%} on Video-LLaVA and \textbf{6\%} on Chat-UniVi, while ICL improved performance by \textbf{6\%} and \textbf{1.5\%}, respectively. However, although the model's performance has improved, the score doesn't increase as significantly as observed in other tasks (for instance, the best-performing fine-tuned model still does not surpass LongVA or Mini-InternVL). We'll illustrate this point in the next section.

\subsection{Analysis}

We perform analysis and ablation studies to further demonstrate the superiority of our metric over traditional metrics. Apart from that, we discuss the possible factors that influence the zero-shot and fine-tuned models' performance. Additional experiments and discussions are shown in the supplementary material.

\subsubsection{Metric Validation}
\label{subsubsec:quanti_ana}

\begin{figure}[htbp] 
    \centering 
    \begin{minipage}{0.8\textwidth} 
        \centering 
        \includegraphics[width=\textwidth]{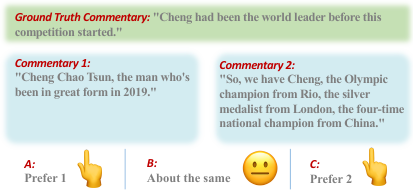} 
    \end{minipage}
    \vspace{1em} 
    \vspace{-2mm}
    \caption{An example question and choices from the user study}
    \label{Fig.userstudy} 
    \vspace{-3mm}
\end{figure}

\begin{figure}[htbp] 
    \centering 
    \begin{minipage}{0.8\textwidth} 
        \centering 
        \includegraphics[width=\textwidth]{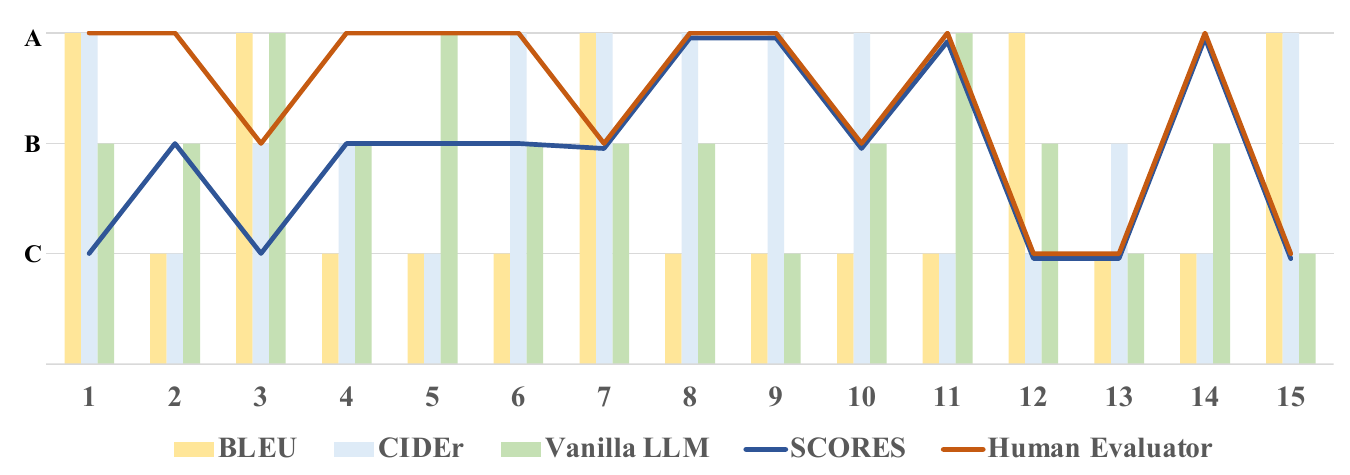} 
    \end{minipage}
    \vspace{1em} 
    \vspace{-2mm}
    \caption{\textbf{Results of user study (15 samples)}. Bars represent traditional metrics (BLEU, CIDEr, Vanilla LLM), while the line represents our SCORES and human evaluators (the dark red part is the overlap).}
    \label{Fig.userstudyres} 
    \vspace{-5mm}
\end{figure}

\textbf{Comparison with Traditional Metric}: In Sec.~\ref{eval&metric}, we theoretically demonstrate the necessity of proposing a new metric for sports commentary tasks. We further conduct experiments using BLEU and CIDEr to evaluate different models, as shown in Tab.~\ref{exp}.\\
\noindent The results indicate that traditional metrics yield low scores across all models (BLEU scores below 1, CIDEr scores around 0.1), making it challenging to distinguish model performance. In contrast, SCORES offers a more nuanced evaluation, effectively and distinctively measuring the model's capabilities. For example, although InterVL-Chat-2 has the lowest BLEU score and a relatively low CIDEr score, our metric highlights its superior performance across all sports categories.\\
\noindent \textbf{Results of Human Evaluation}: To better validate the advantages of our metric, we recruited a total of \textbf{36} human evaluators to conduct our user study. The evaluators were shown the ground truth commentary and two commentaries generated by different models (commentary 1 and commentary 2). Evaluators give a choice for each sample, with A representing a preference for \textit{commentary 1}, B indicating a tie, and C representing a preference for \textit{commentary 2}. We took the most frequent choice as the final result for each sample. Similarly, for other metrics, the results were also categorized into three choices: A, B, and C by comparing the scores, as shown in Fig.~\ref{Fig.userstudy}. \\
\noindent We compare the human-produced results with SCORES and three previous metrics (Vanilla LLM, BLEU, and CIDEr), as shown in Fig.~\ref{Fig.userstudyres}. The results show that the overlap rate between our metric and human decisions is \textbf{60\%}, about \textbf{1.5x} higher than the overlap rate of the best previous method, Vanilla LLM. Furthermore, the overlap rate of Vanilla LLM's results is very close to that of CIDEr, indicating that LLM itself does not perform better than traditional NLP metrics. It is the improvement brought by our proposed SCORES to LLMs that enhances its performance in evaluating commentaries.

\subsubsection{Zero-shot vs Fine-tune}

It is important to note that our evaluation method is essentially designed to leverage the ground truth commentary and the definitions of SCORES to enable our GPT judge to assess the accuracy and completeness of key information in the model-generated commentary, while evaluating the stylistic coherence of the model's commentary with the ground truth, thus yielding an overall score. \\
\noindent Grounded on this, we can interpret many of our experimental results. For zero-shot models, we observed that InternVL-2 achieved significantly higher scores than other models. Upon examining its generated commentaries, we find out that its outputs were notably longer, indicating that it captures more concise and accurate key visual or background information. Apart from this, in the case of fine-tuning, the two methods we proposed primarily enable models to comprehend the stylistic elements of the ground truth commentary and the specific content in SCORES. However, these methods do not enhance the models' ability to capture pivotal visual or background information, which explains the reason for the fine-tuned models' underperformance compared with zero-shot models. Therefore, we believe that establishing a keyword repository for the model to perform Retrieval-Augmented Generation (RAG)~\cite{lewis2021retrievalaugmentedgenerationknowledgeintensivenlp} or employing methods such as VisCoT~\cite{shao2024visualcotadvancingmultimodal} to enhance the model’s ability to capture visual information would be key strategies for further improving the model’s commentary capabilities.

%% file: sec/6_conclusion.tex
\section{Conclusion \& Limitations}

\label{sec:lim&con}

In this paper, we introduce a novel benchmark named \textbf{SCBench} for Video LLMs, focusing on the complex task of sports video commentary. We develop CommentarySet, a dataset comprising 5,775 annotated clips across six sports categories, and propose a six-dimensional metric, SCORES, specially designed to evaluate the task. Our experiments reveal significant limitations in the competence of current models and fine-tuning methods, providing a foundation for advancing Video LLMs' capabilities in complex visual understanding and text generation.\\

\noindent There are two main limitations. First, we have not yet tested SCBench on closed-source Multimodal Large Language Models (MLLMs) that cannot directly process video data, such as GPT-4V and Gemini-1.5 Pro. In previous works, these MLLMs often achieved SOTA, surpassing Video LLMs. We anticipate that Video LLMs could eventually outperform these MLLMs, and we hope that our work will provide new insights for future research. Besides, our proposed fine-tuning methods currently fail to effectively address the fundamental issues of the current model, highlighting the need for future structural improvements.

%% file: sec/X_suppl.tex
\appendix
% \maketitlesupplementary

% \section{Supplementary Material}

\section{Overview}

The following aspects are included in the supplementary material:

\begin{itemize}
    \item Extended Related Work
    \item Supplementary Experiment Details and Analysis
    \item Additional Information About CommentarySet
        \begin{itemize}
            \item File Structure
            \item Benchmark Comparison
            \item Further Discussion
        \end{itemize}
    \item Extended Limitations
\end{itemize}

\section{Extended Related Work}
\subsection{Metrics for Benchmarking VideoLLMs}
In the current benchmarks of Video LLMs, many different metrics have been used. Therefore, we add this section to discuss the metrics employed in these works.

\noindent  In terms of evaluation metrics, most benchmarks, have used QA tasks for assessment. These studies involve extensive manual annotation of videos to create QA pairs. There are two exceptions: one is Youku-mPLUG~\citet{xu2023youkumplug10millionlargescale}, and the other is TempCompass~\citet{liu2024tempcompassvideollmsreally}. Youku-mPLUG generates captions for given videos, which are evaluated using traditional NLP metrics like BLEU, METEOR, and ROUGE. Conversely, TempCompass evaluates MLLMs by curating videos with candidate caption components, requiring VideoLLMs to select appropriate textual elements and generate captions. Accuracy is assessed by backtracking from the generated captions to the originally chosen components with GPT. The above metrics are based on the explicit feature or QA pairs, which fall short of the ability to evaluate the sports commentary task with higher openness and more implicit meaning. In this case, there is an urgent need for a task-specific metric. That's the importance of our proposed SCORES.

\section{Supplementary Experiment Details}
\label{A.sup_exp}

\subsection{Hyperparameters \& Prompts}

For more details of our experiment, we provide additional hyperparameters adopted during the experimental process, as shown in Tab.~\ref{tab:setting}. Furthermore, we also provide the specific content of the prompt/system messages used at each step during our experiments, as detailed below:\\
First, we define the \textbf{Dimension Bank}, which includes the description of the six dimensions, and will be used repeatedly later:\\
\textbf{Dimension Bank:}\\
\textbf{1.} "Key Events Caption: Describe key events in the match, such as scores, turnovers, fouls, substitution etc.",\\
\textbf{2.} "Technical Detail Analysis: Explain the technical actions of the players, such as passing, shooting, defensive strategies, dribbling techniques etc.",\\
\textbf{3.} "Background Information Interpretation: Introduce the background, performance, and individual characteristics of the players or the history, performance records, and current competitive status of the teams.",\\
\textbf{4.} "Tactical Analysis: Explain the tactical arrangements and substitutions of the teams or analyze the execution of tactics by players and teams during the match.",\\
\textbf{5.} "Match Situation Interpretation: Analyze and interpret the current situation on the field, such as the score, trends, etc.; or predict the progression of the match.",\\
\textbf{6.} "Emotional Expression: Convey emotions through interjections or tone, such as 'Great shot!', 'Beautiful!', 'Oh no!', etc., or use vivid language and emotional delivery to enhance the audience's sense of involvement and engagement."

\noindent \textbf{GPT merging}
\begin{lstlisting}[breaklines=true, breakatwhitespace=true, linewidth=\textwidth, language=TeX, numbers = none]
    System Message: "Sports commentary refers to the practice of providing live or post-event verbal descriptions, analyses, and insights about a sports competition by professional commentators. It aims to enhance the viewing experience for audiences by offering detailed information, tactical breakdowns, and highlights of the game. Your task is to assist users in determining whether two sequential commentaries should be merged into one based on their RELEVANCE and FLUENCY. Besides, if one commentary is very short, it is likely needed to be merged."
    Prompt: "The current commentary is: [Current Commentary]; the next commentary is: [Next Commentary]. Should they be merged based on their RELEVANCE and FLUENCY? Respond with 'True' or 'False' ONLY."
\end{lstlisting}

\noindent  \textbf{GPT Label Generation}
\begin{lstlisting}[breaklines=true, breakatwhitespace=true, linewidth=\textwidth, language=TeX, numbers = none]
    System Message: "Sports commentary refers to the practice of providing live or post-event verbal descriptions, analyses, and insights about a sports competition by professional commentators. It aims to enhance the viewing experience for audiences by offering detailed information, tactical breakdowns, and highlights of the game. There are 6 main categories of commentary: [Dimension Bank]. Based on the given commentary, your task is help user select the most appropriate category(s) (one or multiple if applicable). For example, if the commentary is about emotional expression, you need to answer '5'; if it fits both real-time description and background information, you need to answer '1, 2'; if it does not fall into any ot the given categories, answer 'None'."
    Prompt: "The given commentary is: [Target Commentary]; what categories (or category) in the instruction fit(s) the commentary best? NO MORE THAN THREE SELECTIONS. Give me the answr directly, NO EXPLANATION."
\end{lstlisting}

\noindent  \textbf{Six-Dimensional GPT-Based Evaluation (SCORES)}
\begin{lstlisting}[breaklines=true, breakatwhitespace=true, linewidth=\textwidth, language=TeX, numbers = none]
    System Message: "Sports commentary refers to the practice of providing live or post-event verbal descriptions, analyses, and insights about a sports competition by professional commentators. It aims to enhance the viewing experience for audiences by offering detailed information, tactical breakdowns, and highlights of the game. There are [k] main categories of commentary: [chose k description from Dimension Bank according to the given label]. Based on the categories mentioned above, your task is to grade the model commentary to refer to the given ground truth commentary. Please grade the model commentary now, and ONLY GIVE ME ONE floating-point number between 0-10 and NOTHING ELSE.(0 means extremely bad and 10 means perfect)"
    Prompt: "Here are the gt commentary and the model commentary: gt commentary: [GT Commentary]; model commentary : [Model Commentary] Please grade the model commentary now, and PLEASE ONLY GIVE me ONE floating-point number between 0-10 and NOTHING ELSE.(0 means extremely bad and 10 means perfect)"
\end{lstlisting}

\noindent  \textbf{Traditional GPT-Based Evaluation}
\begin{lstlisting}[breaklines=true, breakatwhitespace=true, linewidth=\textwidth, language=TeX, numbers = none]
    System Message: "Sports commentary refers to the practice of providing live or post-event verbal descriptions, analyses, and insights about a sports competition by professional commentators. It aims to enhance the viewing experience for audiences by offering detailed information, tactical breakdowns, and highlights of the game. Your task is to grade the model commentary referring to the given ground truth commentary, based on the similarity. Please grade the model commentary now, and ONLY GIVE ME ONE floating-point number between 0-10 and NOTHING ELSE.(0 means extremely bad and 10 means perfect)"
    Prompt: "Here are the gt commentary and the model commentary: gt commentary: [GT Commentary]; model commentary : [Model Commentary] Please grade the model commentary now, based on the similarity with gt commentary, and PLEASE ONLY GIVE me ONE floating-point number between 0-10 and NOTHING ELSE.(0 means extremely bad and 10 means perfect)"
\end{lstlisting}

\noindent  \textbf{Video LLMs Inference \& ICL Training}
\begin{lstlisting}[breaklines=true, breakatwhitespace=true, linewidth=\textwidth, language=TeX, numbers = none]
    Prompt: "Sports commentary refers to the practice of providing live or post-event verbal descriptions, analyses, and insights about a sports competition by professional commentators. It aims to enhance the viewing experience for audiences by offering detailed information, tactical breakdowns, and highlights of the game. In this video, you should provide commentary based on the [k] categories mentioned below: [chose k description from Dimension Bank according to the given label]. IGNORE the commentator in the video COMPLETELY, which means your commentary should NOT include the commentator. Your task as a professional commentator is to provide commentary to enhance the viewer's understanding and enjoyment of the game. You should provide commentary based on the above mentioned several categories."
\end{lstlisting}

\noindent  \textbf{CoT Training}
\begin{lstlisting}[breaklines=true, breakatwhitespace=true, linewidth=\textwidth, language=TeX, numbers = none]
    Prompt: Sports commentary refers to the practice of providing live or post-event verbal descriptions, analyses, and insights about a sports competition by professional commentators. It aims to enhance the viewing experience for audiences by offering detailed information, tactical breakdowns, and highlights of the game. There are six possible perspectives you should focus on: (1) Key Events Caption: Describes crucial match events like scores, fouls, and substitutions. (2) Technical Detail Analysis: Focuses on player techniques and strategies such as passing, shooting, and defending. (3) Background Information Interpretation: Provides insights into player backgrounds and team histories, including performance records and current standings. (4) Tactical Analysis: Analyzes team tactics and player executions during the match. (5) Match Situation Interpretation: Discusses current match conditions and potential future developments. (6) Emotional Expression: Captures and conveys the emotional aspect of the game with interjections or emotive language to engage the audience further. Based on this sports video and the commentary perspectives mentioned above, provide a concise and coherent commentary.
\end{lstlisting}

\begin{table*}
    \centering
    \caption{This section provides a detailed explanation of the hyperparameters used by the model. \textbf{Others} refers to certain hyperparameters specific to the model. The \textbf{-} symbol does not necessarily indicate that the model lacks this hyperparameter; rather, it means that the parameter is not explicitly set in the configuration.}
    \vspace{-2mm}
    \resizebox{0.95\textwidth}{!}{\begin{tabular}{l|cccc}
    \toprule
        Video LLms &  Max New Tokens &  Temperature & Num Sampled Frames & Others\\
    \midrule
        VILA &  512 &  0.2 &  - &  Num Beams = 1 \\ 
        LLaVA-NeXT-Video &  - &  - &  32 &  Spatial Pooling Stride = 2, Spatial Pooling Output Channels = 1024 \\ 
        Video-LLaVA & 512 &  0.2 &  - &  -  \\ 
        Kangaroo &  512 &  0.2 &  - &  Top-p = 0.9 \\ 
        LongVA &  - &  - &  32 &  32 \\
        Chat-UniVi-1.5 & - & 0.2 & 256(max) & num beams = 1, Top-p = none \\
        Intern-VL-chat-2 &  - &  - & - &  Num Segments = 32, Input Size = 448\\
        Intern-VL-Chat-1.5 &  - &  - &  - & Num Segments = 32, Input Size = 448 \\
    \bottomrule
    \end{tabular}}
    \label{tab:setting}
\end{table*}

\subsection{Additional Experiment Analysis}

\subsection{Comparison with Traditional Metrics}

In the Analysis section of our experiment, we examined the advantages of our Metric compared to traditional NLP Metrics such as BLEU and CIDEr. However, we have not yet conducted a comparison with GPT evaluation without our proposed six-dimensions (A.K.A Vanilla GPT Evaluation). \\
\noindent The process of Vanilla GPT Evaluation is as follows: First, we provide the definition of sports video commentary task to the GPT judge. Then, for each video clip, we input both the ground truth commentary and the model-generated commentary. Finally, we ask the GPT to directly score the model-generated commentary based on its similarity to the ground truth. The detailed prompt is mentioned above.\\
\noindent  In Tab.~\ref{abl}, we demonstrate the results of both NLP metrics and Vanilla GPT Evaluation. From the table, we can observe that while Vanilla GPT Evaluation exhibits a certain degree of consistency with our results in terms of model ranking, the scores produced by this method are very close to each other, making it difficult to clearly distinguish the capabilities of different models. Moreover, this method does not offer the possibility for more granular evaluation and analysis. In contrast, our approach not only assesses the overall performance of the models but also evaluates specific sub-abilities, such as the ability to express emotions, as demonstrated by the results under Label-6(Emotional Expression).

\subsubsection{Analysis Cross Different Sports}

\begin{figure*}[htbp]
    \centering
    \includegraphics[width=0.9\textwidth]{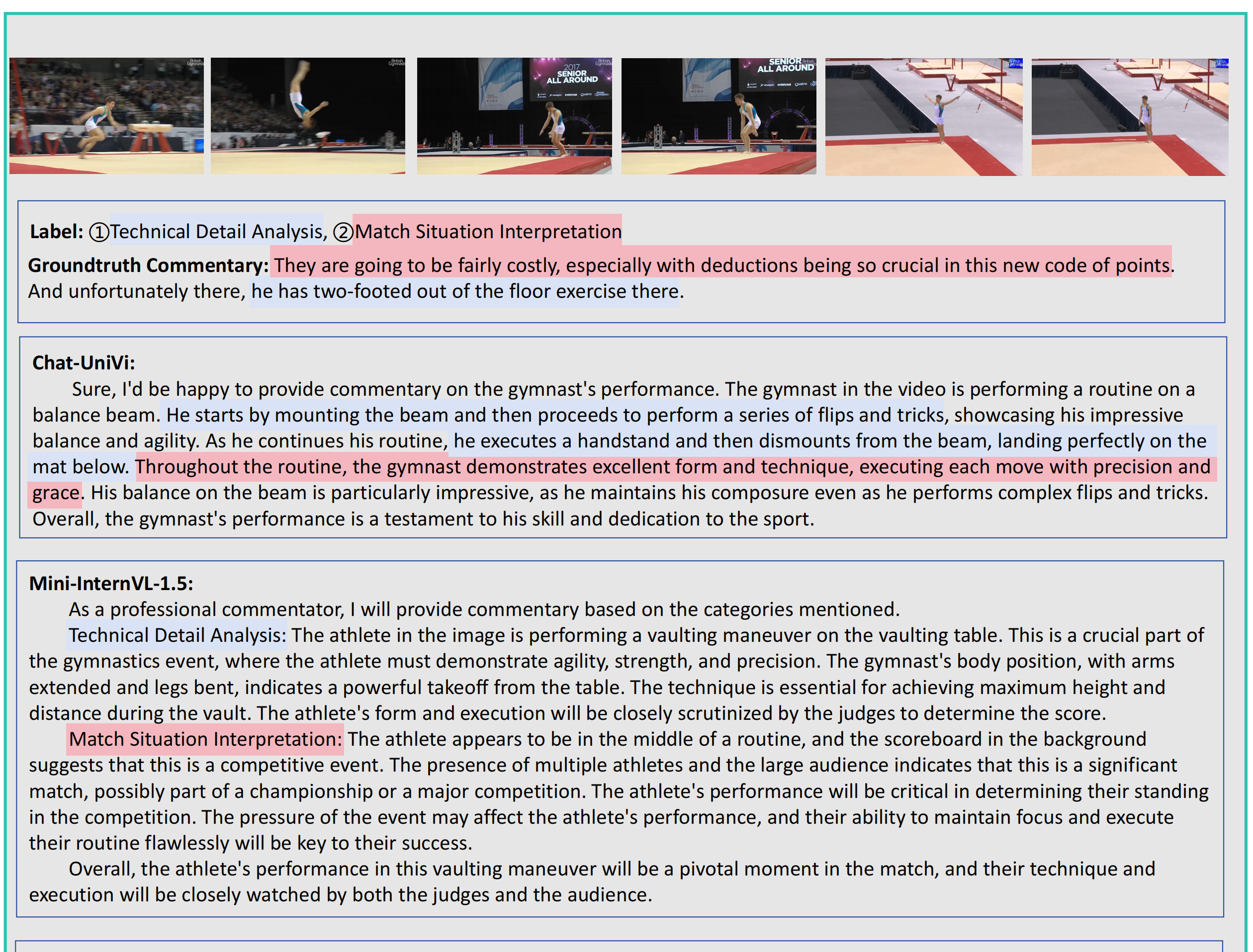}
    \caption{\textbf{An example of the comparison of the generated commentaries in tabletennis.}}
    \label{fig:sample_gym}
\end{figure*}

The characteristics of each sport, including complexity, action frequency, and game pace, significantly influence how models perform in generating commentary\\.
\noindent Sports requiring detailed commentary, such as gymnastics and tennis, which have an average duration of 16.49 and 26.27 seconds with a commentary length of 34.68 and 27.86 words respectively, challenge models to maintain coherence over longer durations. Models like InternVL-Chat-2 excel in these sports, scoring \textbf{5.31} in gymnastics and \textbf{4.98} in tennis, demonstrating superior performance in processing long videos and generating extended commentary texts.\\
\noindent In fast-paced sports like table tennis and basketball which have shorter durations and higher commentary frequency, the model's ability to process and comprehend dense events is challenged. LLaVA-NeXT-Video performs better in these two categories than tennis and gymnastics, indicating its excellence in comprehending dense events. Here we list more examples of the model-generated commentaries across six categories of sports in Fig.~\ref{fig:sample_bask},~\ref{fig:sample_ath},~\ref{fig:sample_soc},~\ref{fig:sample_ten},~\ref{fig:sample_tab},~\ref{fig:sample_gym},.

\begin{figure*}[htbp]
    \centering
    \begin{subfigure}[b]{0.78\textwidth} 
        \centering
        \includegraphics[width=\textwidth]{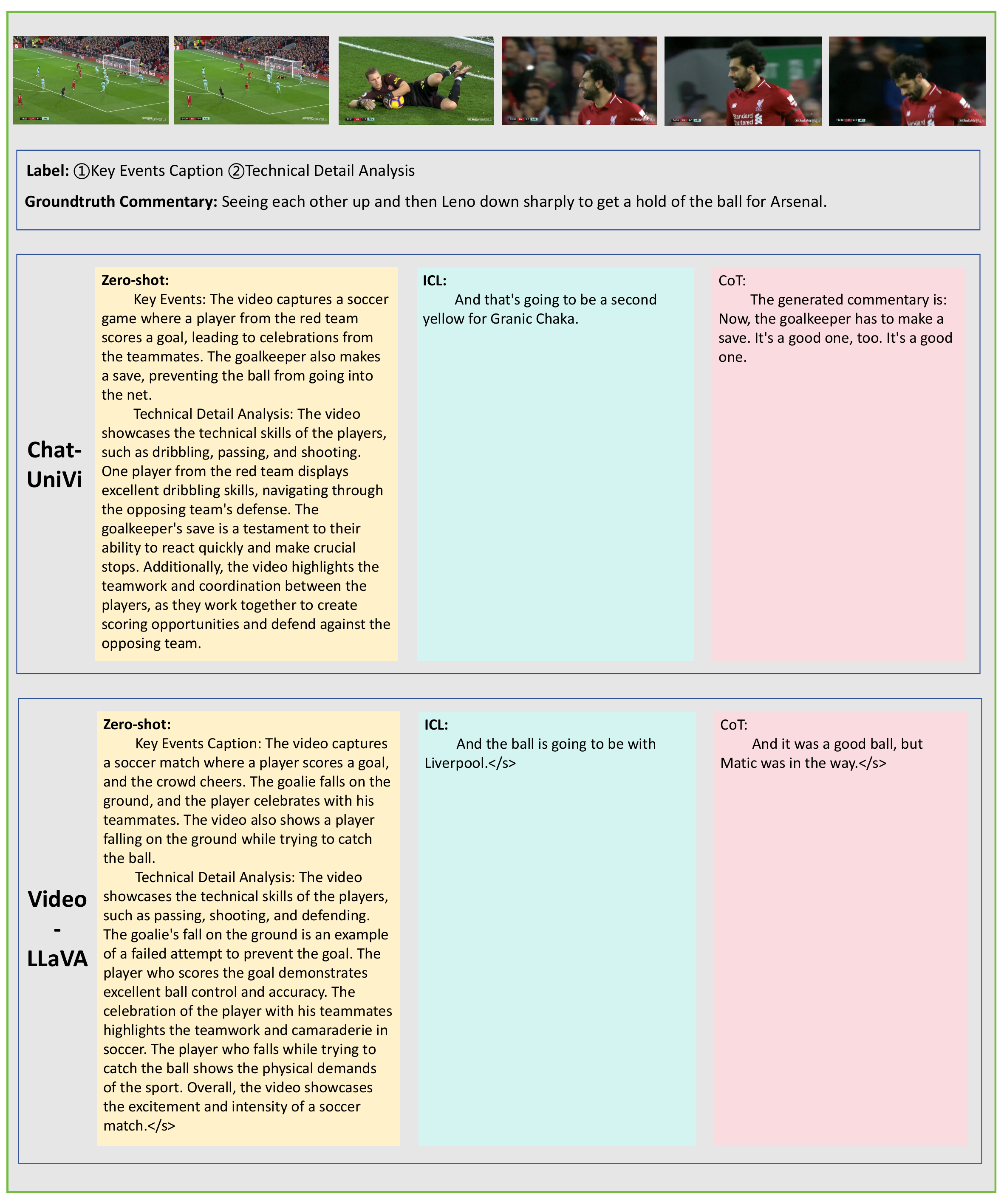} 
    \end{subfigure}
    
    \vspace{-0.4cm} 
    \begin{subfigure}[b]{0.78\textwidth}
        \centering
        \includegraphics[width=\textwidth]{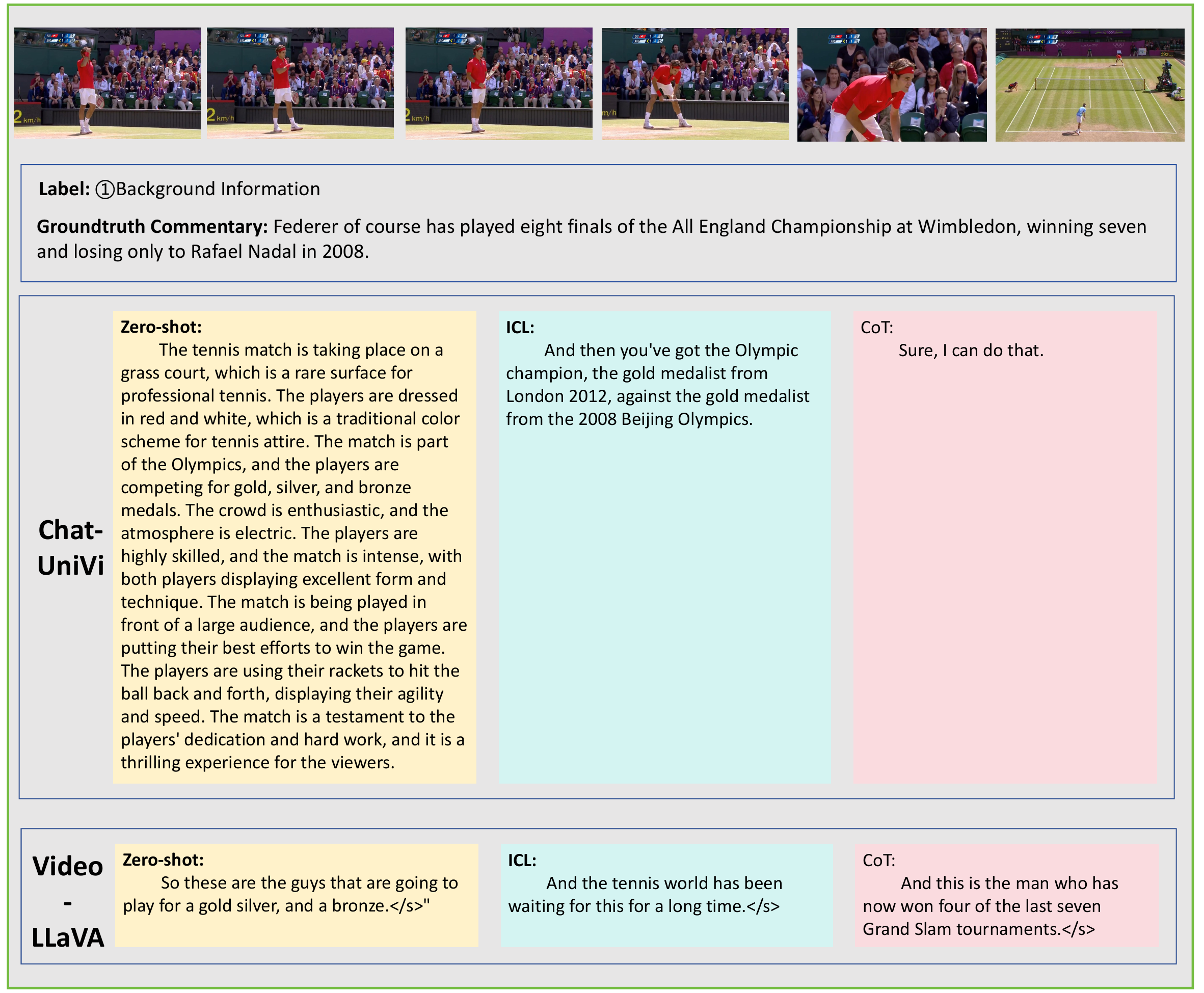} 
    \end{subfigure}
    \caption{\textbf{Examples of the comparison of the generated commentaries by zero-shot and fine-tuned models in soccer and tennis.}}
    \label{fig:main}
\end{figure*}

\subsubsection{Comparison Between Zero-shot \& Fine-tuned Models}

We further present a more detailed comparison of the outputs from models fine-tuned with ICL and CoT training methods versus those of the zero-shot model, as shown in Fig.~\ref{fig:main}. As mentioned in the experimental section of the paper, the outputs from models fine-tuned using these two methods demonstrate a noticeable adoption of specific commentary styles compared to the zero-shot models. This is evident from the increased volume of specific content and reduced redundancy in the generated commentary. Additionally, the fine-tuned outputs are more detailed rather than general captions. However, compared to the ground-truth commentary, the fine-tuned outputs still exhibit issues such as inaccurate or incorrect information representation, and in some cases, even hallucinations (see the result of ICL in the upper part of Fig.~\ref{fig:main}). These observations indicate that further improvements at the structural level of the model are necessary to address these issues fundamentally, thereby enhancing the model's ability to generate accurate and context-aware sports video commentary.

\begin{table*}
    \centering
    \caption{The comparison of various benchmarks encompasses several key aspects: the number of clips (\textbf{Clips}), the average duration of the videos (\textbf{Average Duration}), the method of annotation (\textbf{Anno.}, M/A means the manually/automatic manner), the average number of prompt tokens (\textbf{Prompt Tokens}), whether the videos are sourced from a broad range of domains (\textbf{Multi-Domain}), and whether provide task specific multi-dimensional labels for the commentary (\textbf{Multi-Dimensional Label}).}
    \vspace{-2mm}
    \resizebox{0.95\textwidth}{!}{
    \begin{tabular}{l|cccccc}
    \toprule
        Dataset &  Clips &  Average Duration(s) &  Anno. & Prompt Tokens & Multi-Domain & Multi-Dimensional Label\\
    \midrule
        MSRVTT-QA~\cite{2017Video} &  2,990 &  15.2 &  A &  8.4 & \CheckmarkBold & \XSolidBrush\\ 
        MSVD-QA~\cite{2017Video} &  504 &  9.8 &  A &  7.6 & \CheckmarkBold & \XSolidBrush\\ 
        TGIF-QA~\cite{jang2017tgifqaspatiotemporalreasoningvisual} &  9,575 &  3.0 &  A\&M &  20.5 & \CheckmarkBold & \XSolidBrush \\ 
        ActivityNet-QA~\cite{yu2019activitynetqadatasetunderstandingcomplex} &  800 &  111.4 &  M &  10.2 & \XSolidBrush & \XSolidBrush\\ 
        NExT-QA~\cite{2021NExT} &  1,000 &  39.5 &  A &  25.3 & \CheckmarkBold & \XSolidBrush\\ \hline
        MVBench~\cite{li2024mvbenchcomprehensivemultimodalvideo} &  3,641 &  16.0 &  A &  27.3 & \CheckmarkBold & \XSolidBrush\\
        Video-Bench~\cite{ning2023video} &  5,917 &  56.0 &  A\&M &  21.3 & \CheckmarkBold & \XSolidBrush\\
        EgoSchema~\cite{mangalam2023egoschemadiagnosticbenchmarklongform} &  5,063 &  180.0 &  A\&M & 126.8 & \XSolidBrush & \XSolidBrush \\
        AutoEval-Video~\cite{chen2024autoevalvideoautomaticbenchmarkassessing} &  327 &  14.6 &  M &  11.9 & \CheckmarkBold & \XSolidBrush\\
        TempCompass~\cite{liu2024tempcompassvideollmsreally} &  410 &  11.4 &  A\&M &  49.2 & \CheckmarkBold & \XSolidBrush\\
        Video-MME~\cite{fu2024videommefirstevercomprehensiveevaluation} &  900 &  1017.9 &  M &  35.7 & \CheckmarkBold & \XSolidBrush\\ 
        \hline
        SCBench (ours)&  5,775 &  14.5 &  A\&M & 207.9 &\CheckmarkBold & \CheckmarkBold \\
    \bottomrule
    \end{tabular}}
    \label{tab:comparebench}
\end{table*}

\subsection{Samples of the Human Evaluation Questionaire}

\begin{figure*}[htbp]
    \centering
    \includegraphics[width=0.8\textwidth]{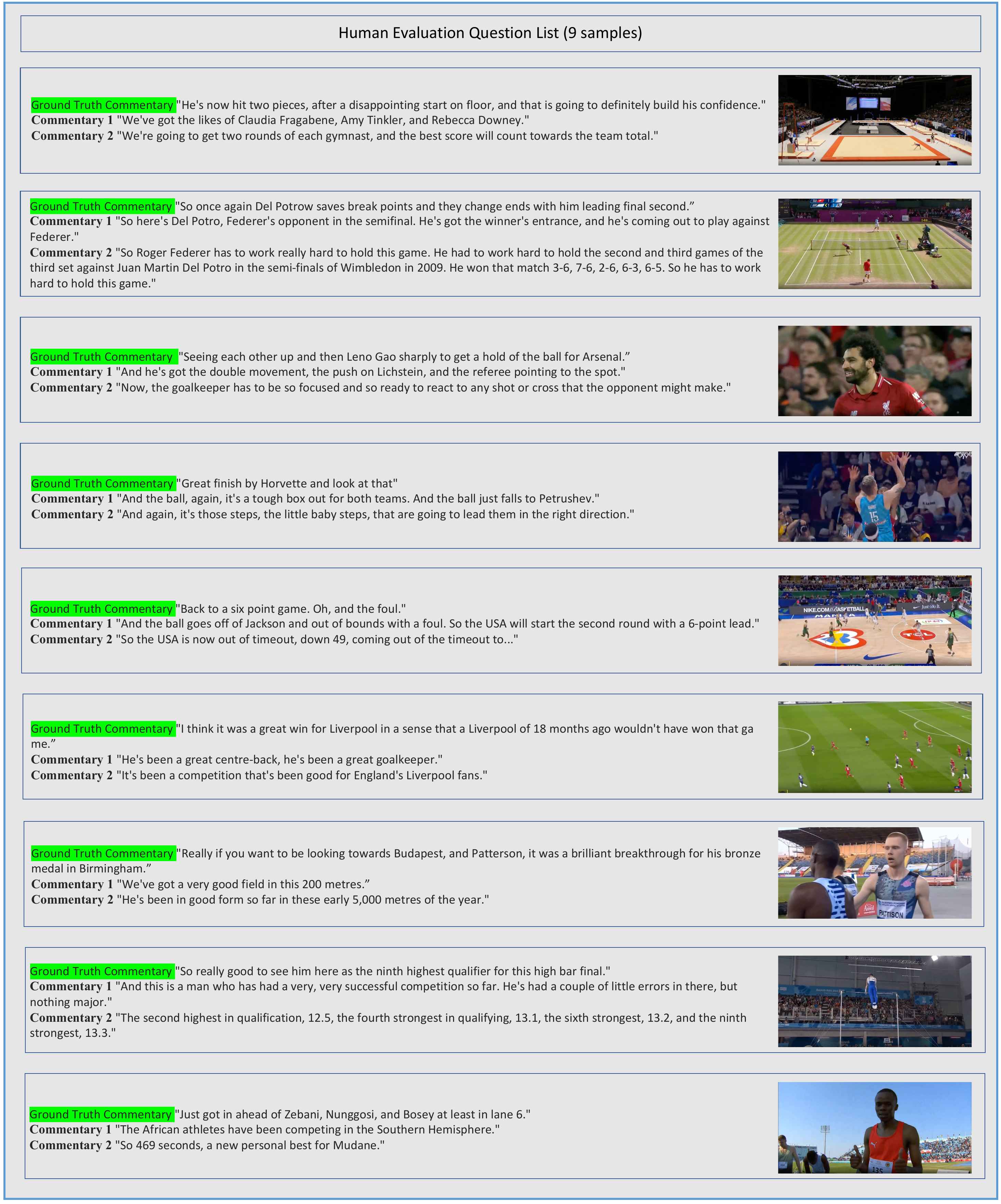}
    \caption{Samples of our Human Evaluation Questionnaire }
    \label{fig:human}
\end{figure*}

Here, we list and visualize some of the questions from the human evaluation questionnaire, including the ground truth commentary, commentary 1, and commentary 2, as shown in Fig.~\ref{fig:human}.

\section{Additional Information About CommentarySet}
\label{A.4}
\subsection{File Structure} In this section, we introduce the structure of CommentarySet:\\
\begin{itemize}
    \item \textbf{CommentarySet}
    \begin{itemize}
        \item \textbf{commentary}
        \begin{itemize}
            \item \texttt{athletics\_final.json}
            \item \texttt{basketball\_final.json}
            \item \ldots
        \end{itemize}
        \item \textbf{video}
        \begin{itemize}
            \item \textbf{athletics}(category)
            \begin{itemize}
                \item \textbf{001(video\_id)/}\texttt{5(clip\_id).mp4}
                \item \textbf{001(video\_id)/}\texttt{7(clip\_id).mp4}
                \item \ldots
            \end{itemize}
            \item \textbf{basketball}(category)
            \item \ldots
        \end{itemize}
        \item \textbf{test.json}
        \item \textbf{train.json}
    \end{itemize}
\end{itemize}
\noindent File \texttt{test.json} and \texttt{train.json} respectively contain the lists of samples in the test subset and the train subset.
\noindent File \texttt{athletics\_final.json} contains information related to each clip, detailed as follows:
\begin{itemize}
    \item \textbf{id}: The id of the clip, where each clip has a unique id. The structure is \texttt{"video\_id" + "-" + "clip\_id"}. For example, the clip with the id \texttt{"003-992"} in \texttt{athletics\_final.json} represents clip 992 of video 003 in athletics.
    \item \textbf{category}: The sports category to which the video in the clip belongs.
    \item \textbf{timestamp} (\texttt{start\_time} \& \texttt{end\_time}): The start and end timestamps of the clip in the original video.
    \item \textbf{text}: The ground truth commentary corresponding to the clip.
    \item \textbf{label}: The six-dimensional label corresponding to the clip.
\end{itemize}

\subsection{Benchmarks Comparison} 
We compare the key elements of our dataset and benchmark with previous benchmarks, as shown in Tab.~\ref{tab:comparebench}

\subsection{Further Discussin} 
Aside from the detailed description of the dataset provided in the article, in this section, we'll provide more points, including the importance of CommentarySet and its potential applications:
\begin{itemize}
    \item CommentarySet is the first commentary task encompasses a wide range of sports. As mentioned in the article, existing datasets like TenniSet~\cite{Faulkner2017TenniSetAD} and SoccerNet~\cite{deliège2021soccernetv2datasetbenchmarksholistic}, which are sports video caption datasets, mainly focus on a single sport and do not provide commentary. Therefore, CommentarySet is an unprecedented dataset for sports video commentary tasks.
    \item This paper applies the dataset to Video LLMs, and it can also be used to Image LLMs and MLLMs. The task of generating commentary based on a given video and prompt can be transformed into other forms. For example, if we convert the video into a series of images and use ImageLLMs to generate commentary for these consecutive images, it can also be an extended task on CommentarySet. Therefore, CommentarySet supports more multimodal sports video commentary tasks, providing opportunities for future work.
    \item By using the six-dimensional analysis method like CommentarySet, we can conduct more in-depth research on commentary styles. For instance, by statistically analyzing a large amount of commentary material for a particular sport or commentator and calculating the proportions of the six dimensions, we can determine the tendencies in their commentary style In this way, we might be able to mimic various commentary styles by pre-specifying the six-dimensional ratios for models while generating commentary.
    \item CommentarySet can also be applied in other research areas and tasks, such as human-computer interaction, real-time sports video commentary, and sports game commentary.
\end{itemize}

\begin{table*}
    \centering
    \caption{\textbf{Comparisson with Traditional Metrics} Models' performance across different sub-tasks in CommentarySet evaluated by multiple metrics.}
    \resizebox{\textwidth}{!}{\begin{tabular}{l|c|ccc|c}
    \toprule
    \multirow{2}{*}{\textbf{Model Name}} & \multirow{2}{*}{\begin{tabular}[c]{@{}c@{}}\textbf{Model}\\\textbf{Params}~~\end{tabular}}
    &\multicolumn{3}{c}{\textbf{Traditional Metric}} & \multirow{2}{*}{\textbf{Average On Our Metric(0-10)}} \\
    \cmidrule{3-5}
     & & BLEU(0-100) & CIDEr(0-1) & Vanilla GPT(0-10)\\
    \midrule
    Mini-InternVL-Chat-4B-V1-5 & 4B & 0.43 & \textbf{0.12} & 2.95 & 4.40 \\
    Video-LLaVA & 7B & 0.67 & 0.08 & 2.07  & 2.55 \\
    LongVA & 7B & 0.39 & 0.10 & 2.67 & 3.85 \\
    LLaVA-NeXT-Video & 7B & 0.35 & 0.10 &  2.90 & 4.20 \\
    Chat-UniVi-1.5 & 7B & 0.56 & 0.11 & 2.64 & 3.37 \\
    Kangaroo & 8B & 0.83 & 0.01 & 1.26 & 1.79 \\ 
    VILA & 34B & 0.66 & 0.09 & 2.74 & 3.40 \\ 
    InternVL-Chat-2 & \textbf{40.1B} & 0.32 & 0.11 & \textbf{3.33} & \textbf{5.44} \\ \hline
    Video-LLaVA-f-c & 7B & 2.11 & 0.06 & 2.18 & 2.70 \\
    Chat-UniVi-1.5-c & 7B & \textbf{2.57} & 0.06 & 2.52 & 3.42 \\
    \bottomrule
    \end{tabular}}
    \label{abl}
\end{table*}

\section{Extended Limitations}

Except the limitations mentioned in the paper, our work has the following two limitations: First, although we proposed a feasible data collection method, the dataset we collected is relatively small and does not cover a diverse range of sports categories (currently limited to six types). Future work can expand the CommentarySet using our proposed method to include more sports types. Second, while we introduced two training methods, the commentary capabilities of the current models still lag behind human-level performance. Further research is needed to develop more effective model architectures or training methods to address this gap.

\begin{figure*}[htbp]
    \centering
    \includegraphics[width=0.9\textwidth]{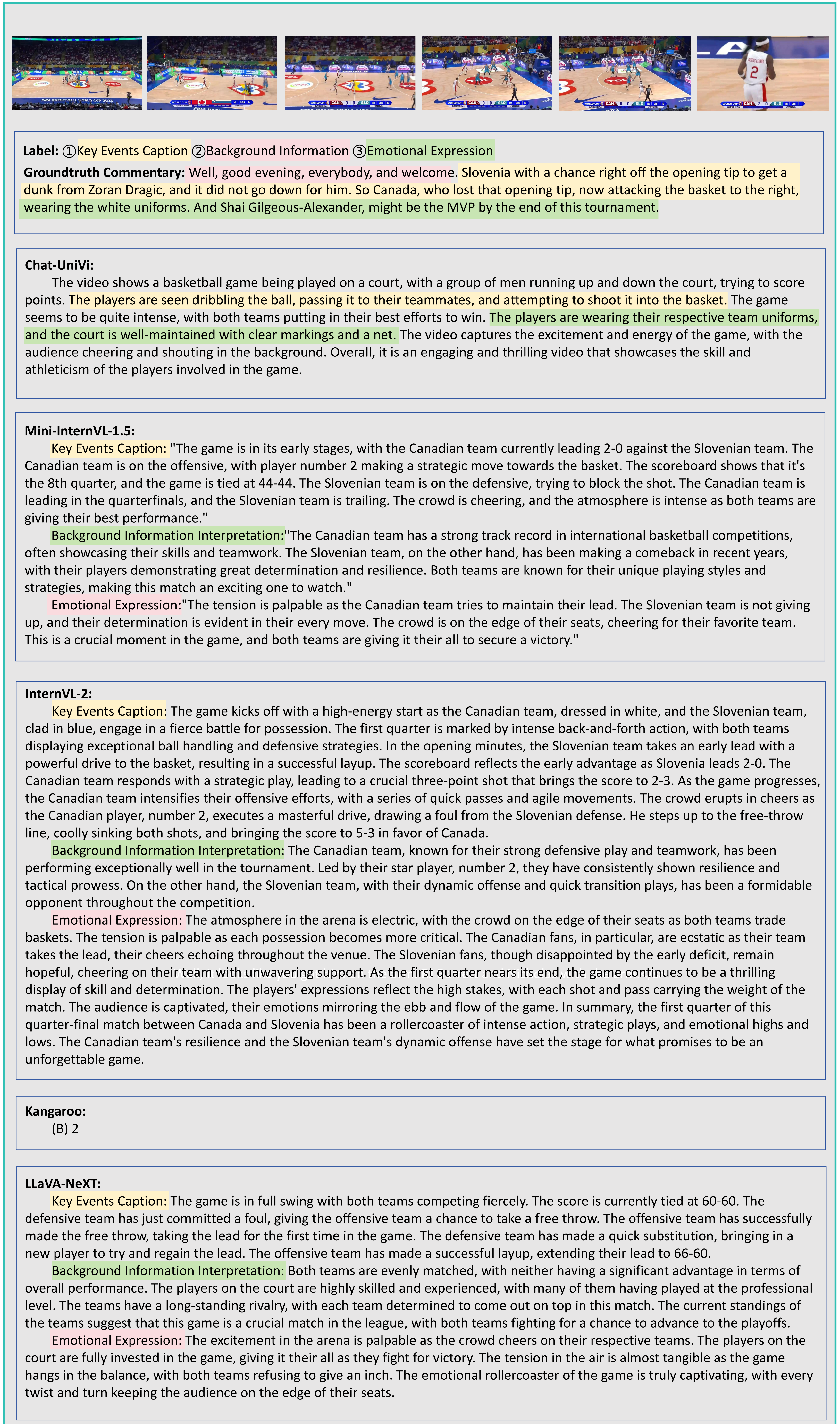}
    \caption{\textbf{An example of the comparison of the generated commentaries in basketball. The sections highlighted in different colors represent the content corresponding to each label. The same applies to subsequent cases.}}
    \label{fig:sample_bask}
\end{figure*}
\begin{figure*}[htbp]
    \ContinuedFloat
    \centering
    \includegraphics[width=0.9\textwidth]{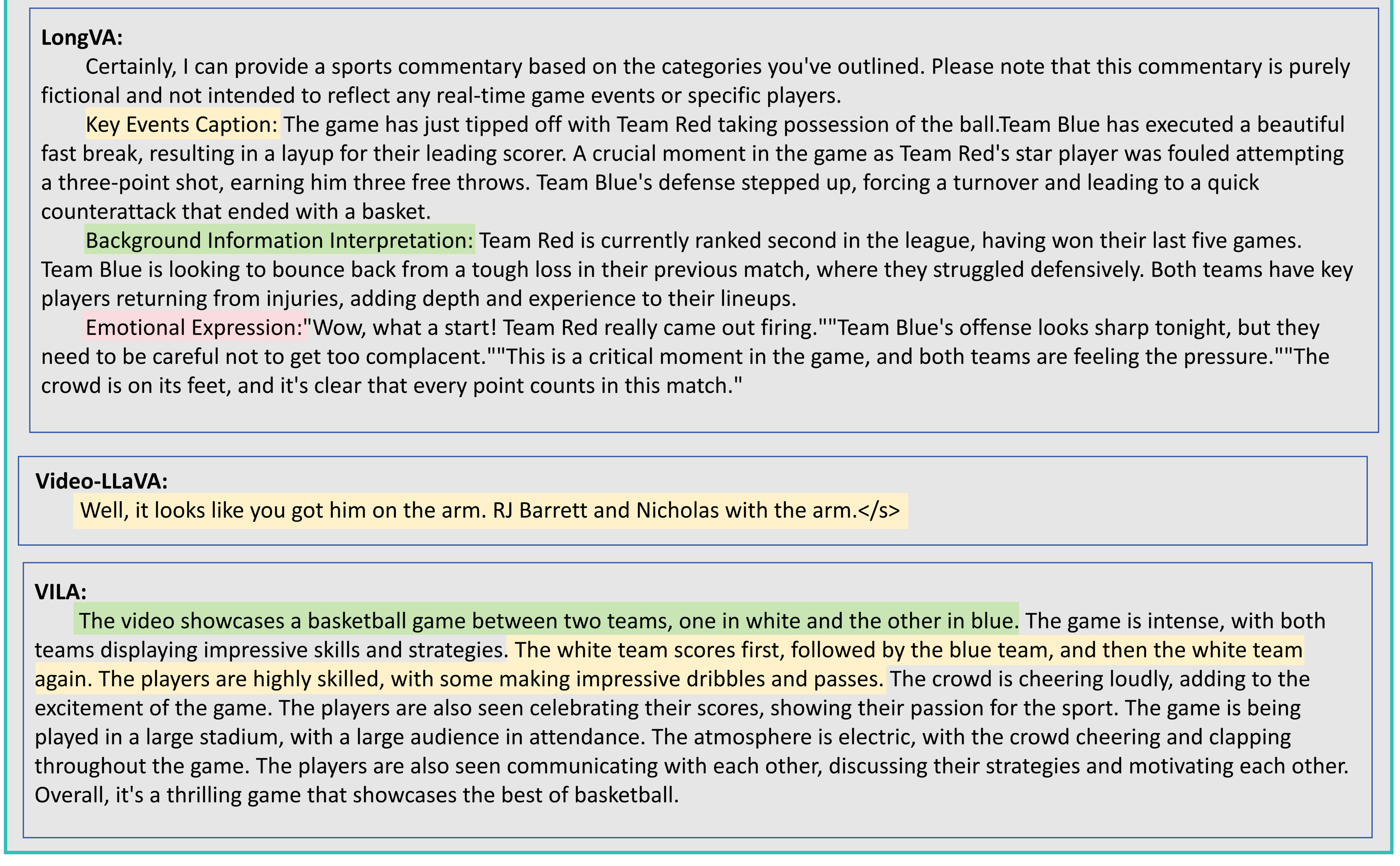}
    \caption{\textbf{An example of the comparison of the generated commentaries in basketball.}}
    \label{fig:sample_bask_continue}
\end{figure*}

\begin{figure*}[htbp]
    \centering
    \includegraphics[width=0.9\textwidth]{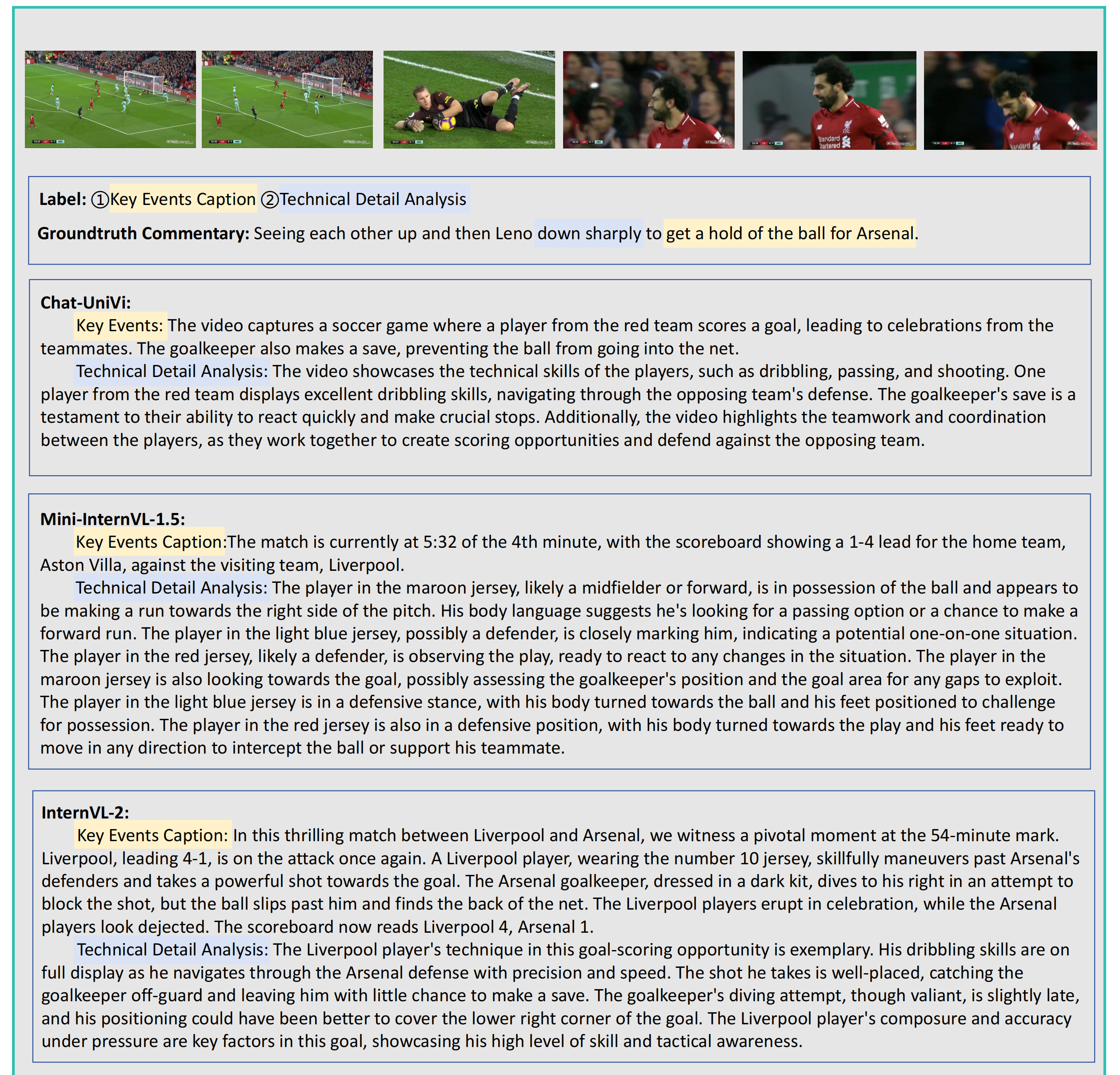}
    \caption{\textbf{An example of the comparison of the generated commentaries in soccer.}}
    \label{fig:sample_soc}
\end{figure*}
\begin{figure*}[htbp]
    \ContinuedFloat
    \centering
    \includegraphics[width=0.9\textwidth]{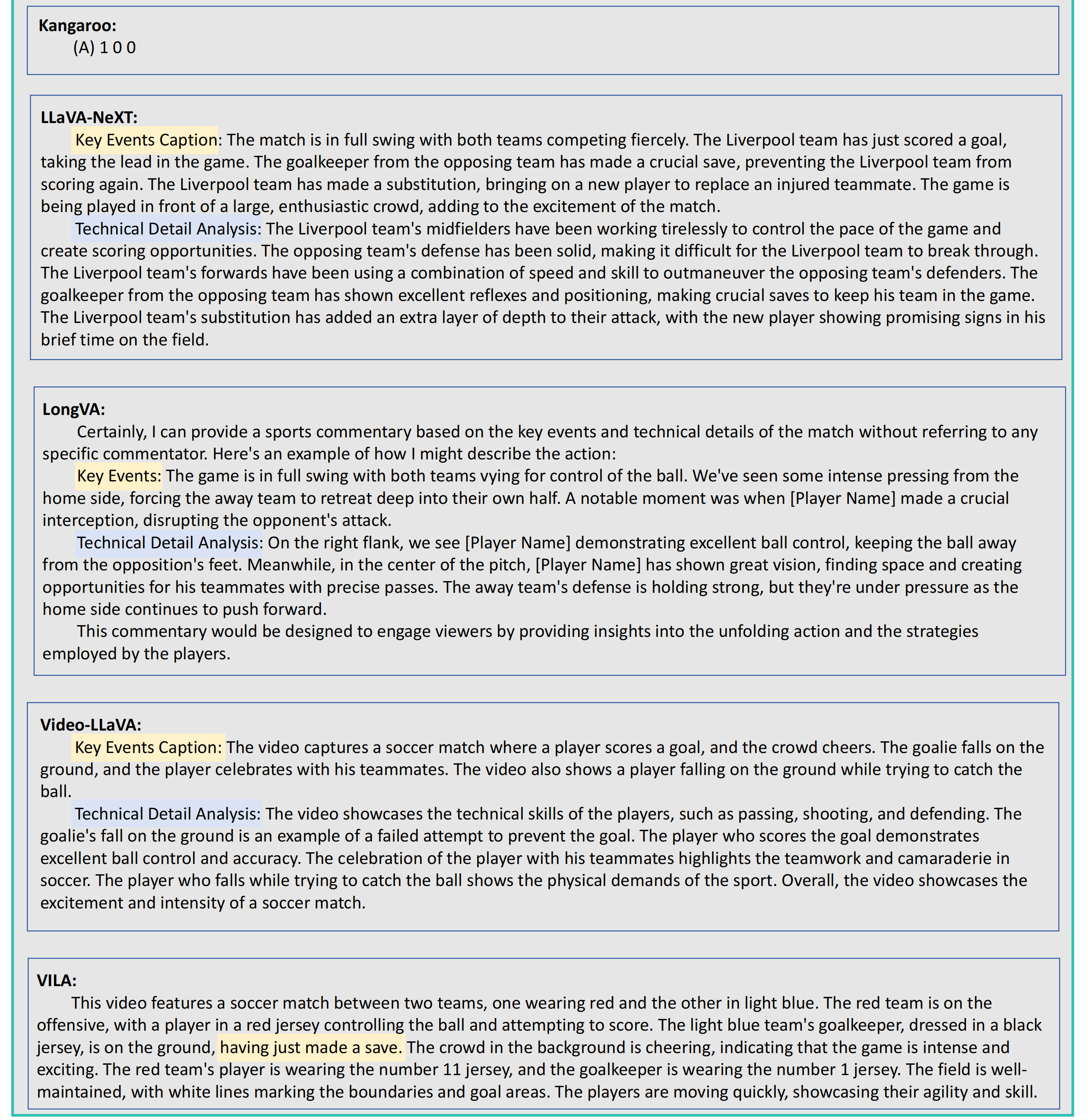}
    \caption{\textbf{An example of the comparison of the generated commentaries in soccer.}}
    \label{fig:sample_soc_}
\end{figure*}

\begin{figure*}[htbp]
    \centering
    \includegraphics[width=0.9\textwidth]{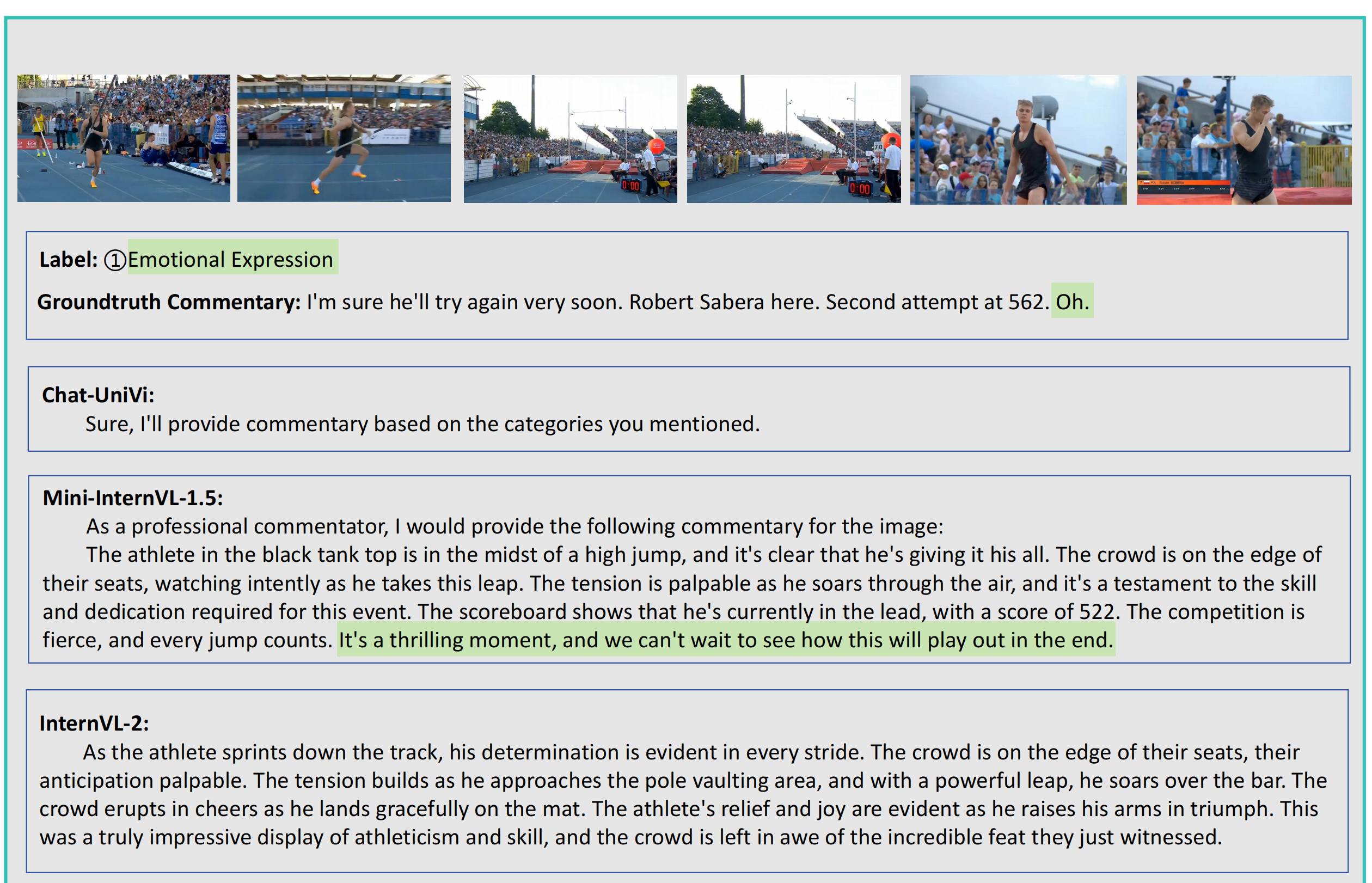}
    \caption{\textbf{An example of the comparison of the generated commentaries in athletics.}}
    \label{fig:sample_ath}
\end{figure*}
\begin{figure*}[htbp]
    \ContinuedFloat
    \centering
    \includegraphics[width=0.9\textwidth]{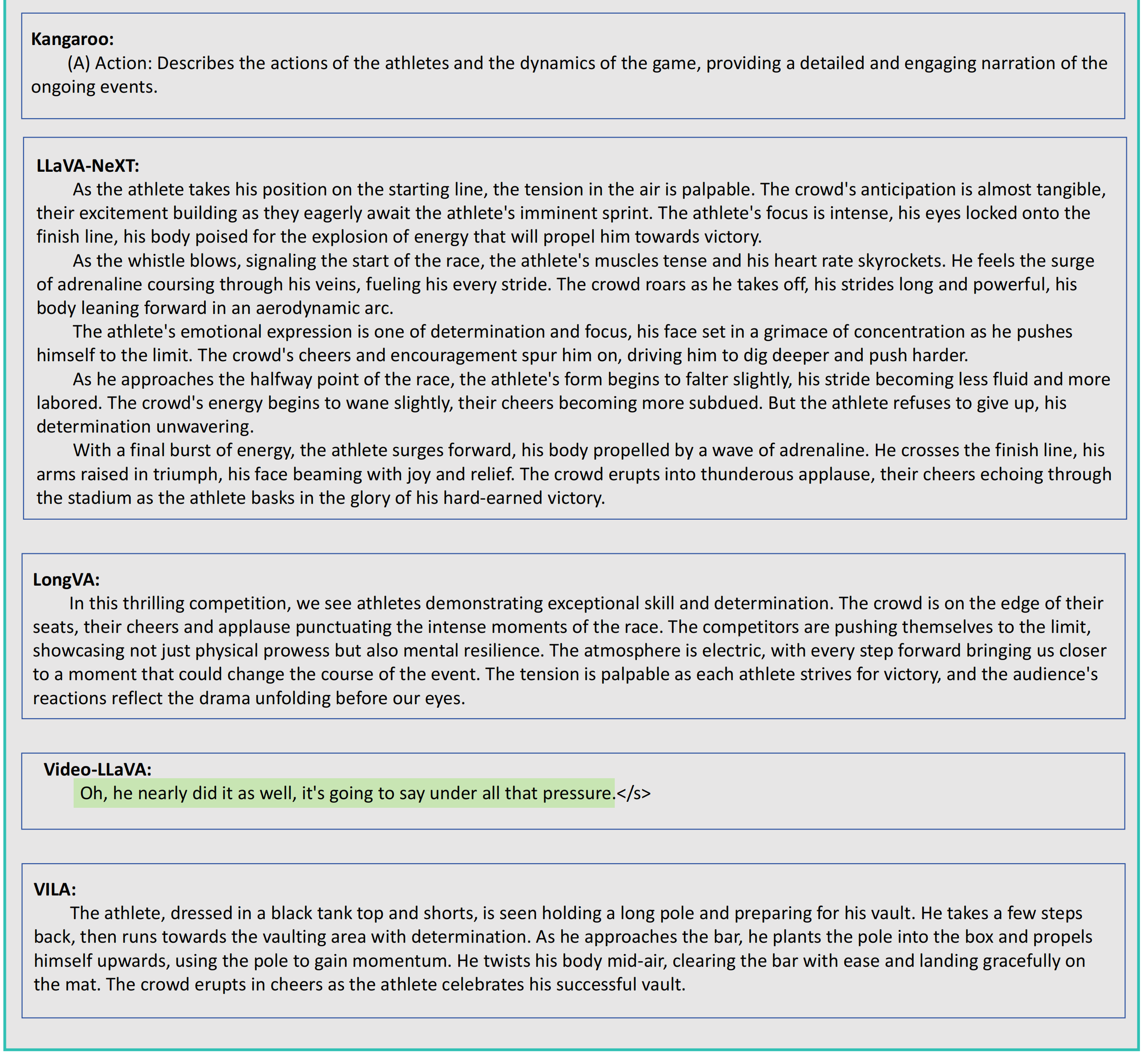}
    \caption{\textbf{An example of the comparison of the generated commentaries in athletics.}}
    \label{fig:sample_ath_}
\end{figure*}

\begin{figure*}[htbp]
    \centering
    \includegraphics[width=0.9\textwidth]{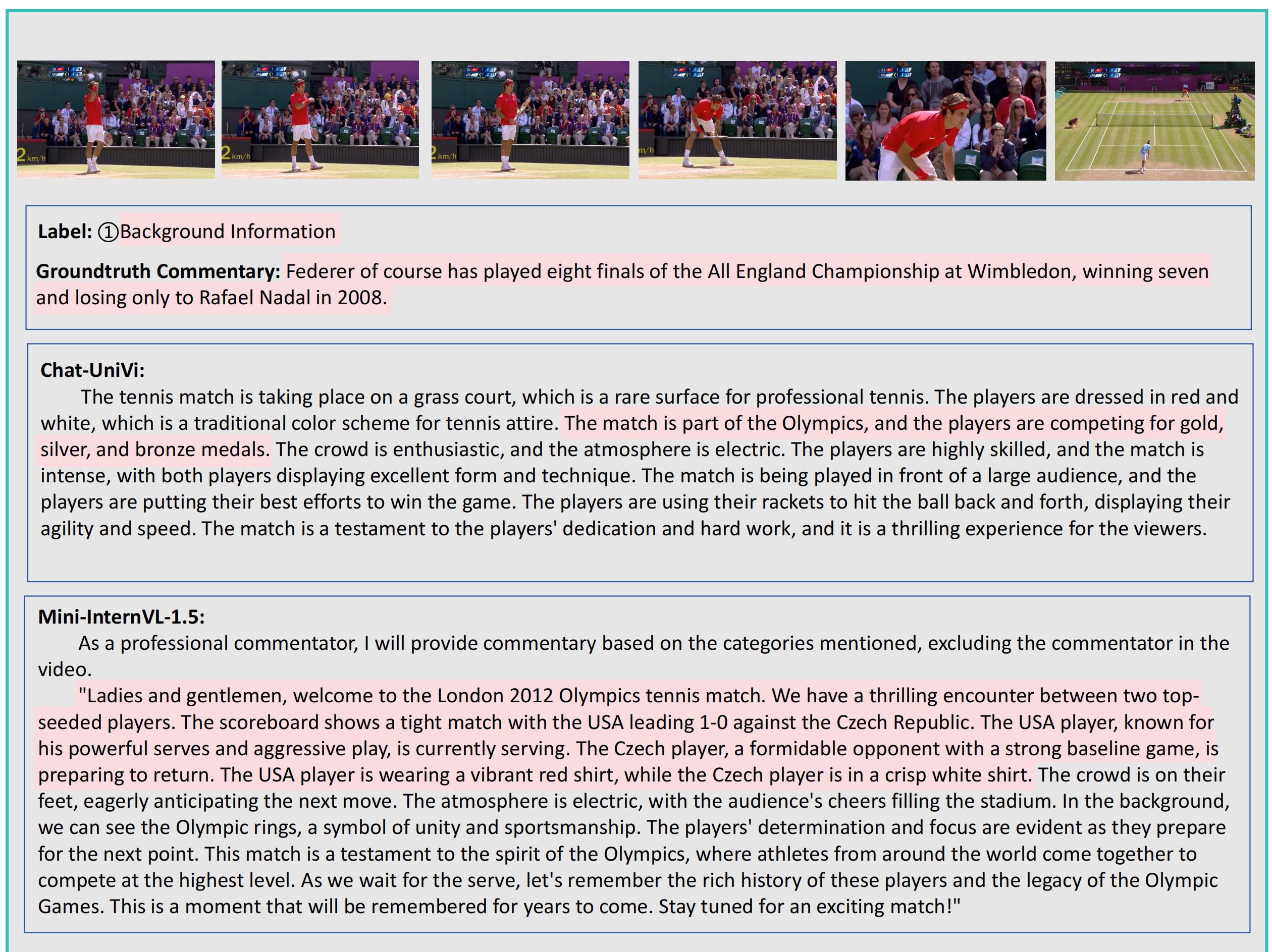}
    \caption{\textbf{An example of the comparison of the generated commentaries in tennis.}}
    \label{fig:sample_ten}
\end{figure*}
\begin{figure*}[htbp]
    \ContinuedFloat
    \centering
    \includegraphics[width=0.9\textwidth]{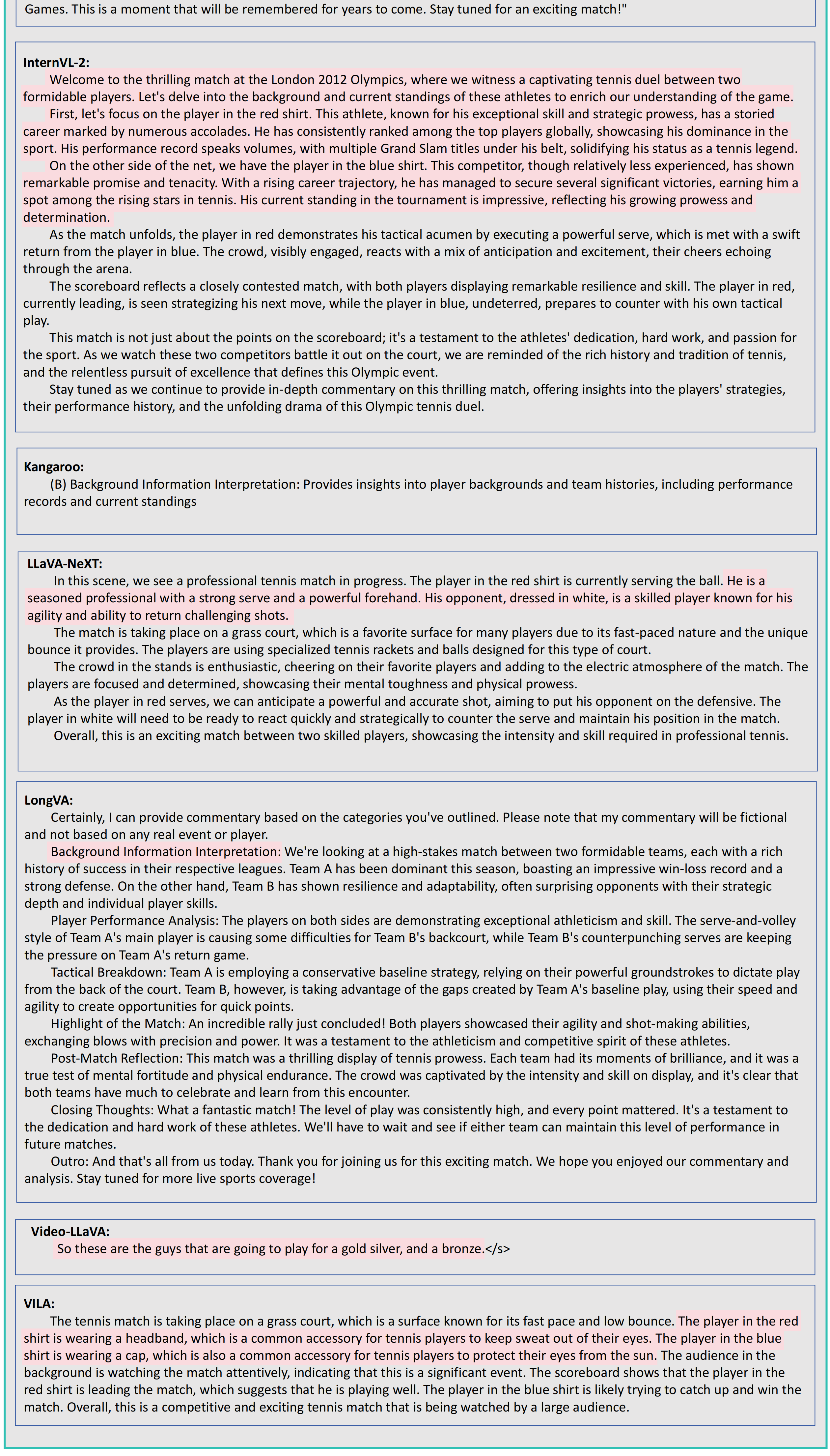}
    \caption{\textbf{An example of the comparison of the generated commentaries in tennis.}}
    \label{fig:sample_ten_}
\end{figure*}

\begin{figure*}[htbp]
    \centering
    \includegraphics[width=0.9\textwidth]{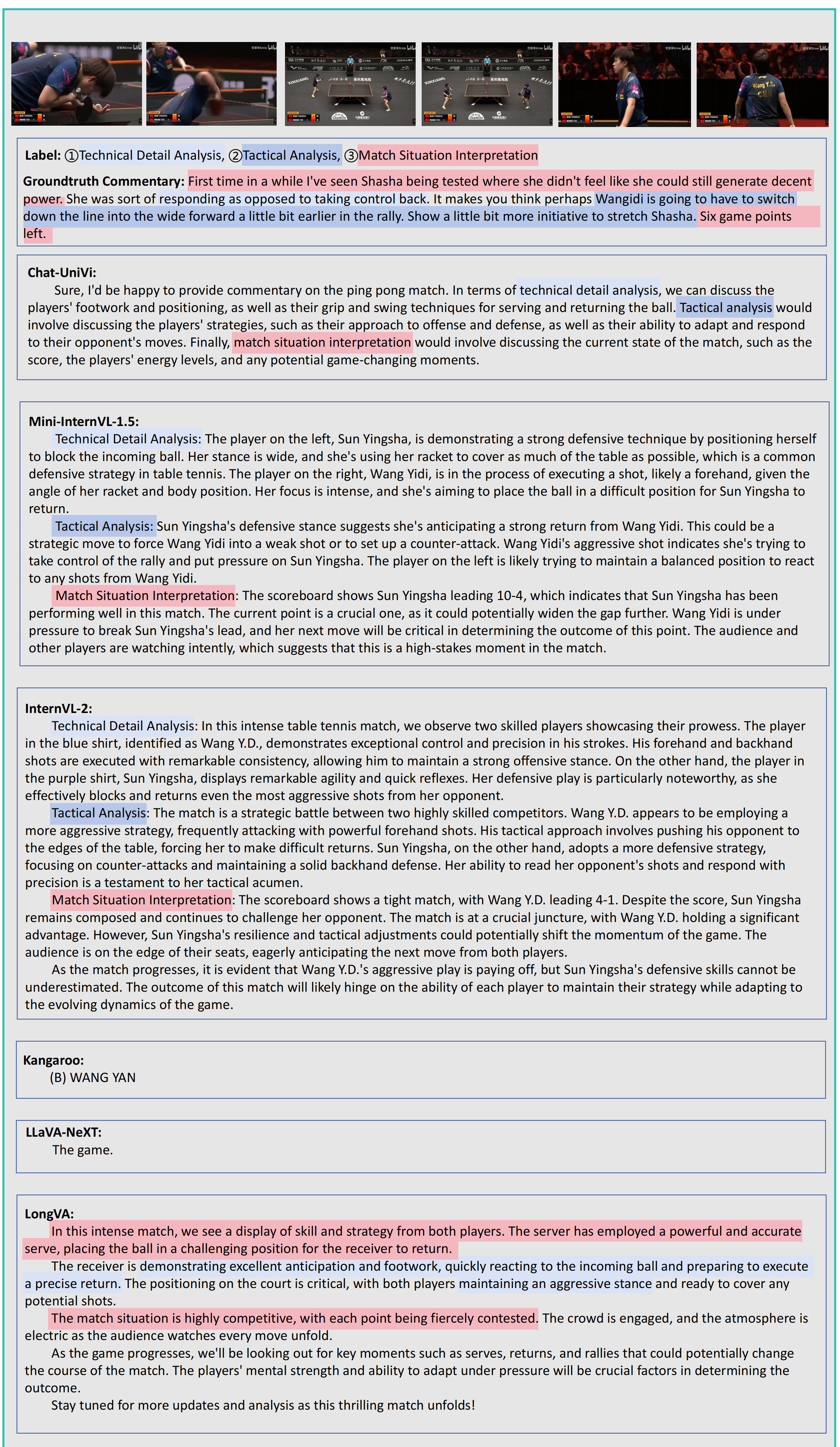}
    \caption{\textbf{An example of the comparison of the generated commentaries in tabletennis.}}
    \label{fig:sample_tab}
\end{figure*}
\begin{figure*}[htbp]
    \ContinuedFloat
    \centering
    \includegraphics[width=0.9\textwidth]{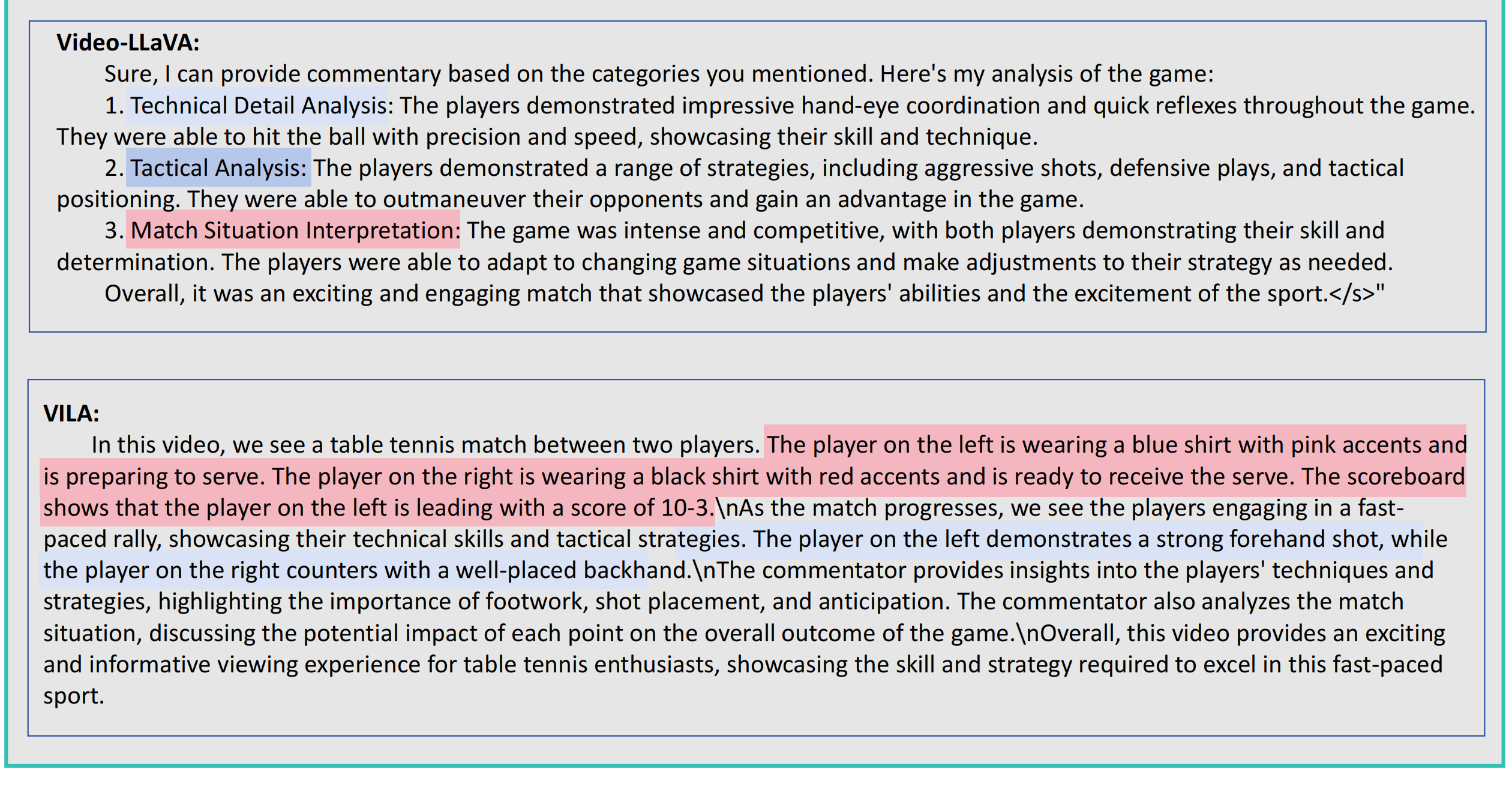}
    \caption{\textbf{An example of the comparison of the generated commentaries.}}
    \label{fig:sample_tab_}
\end{figure*}

\begin{figure*}[htbp]
    \centering
    \includegraphics[width=0.9\textwidth]{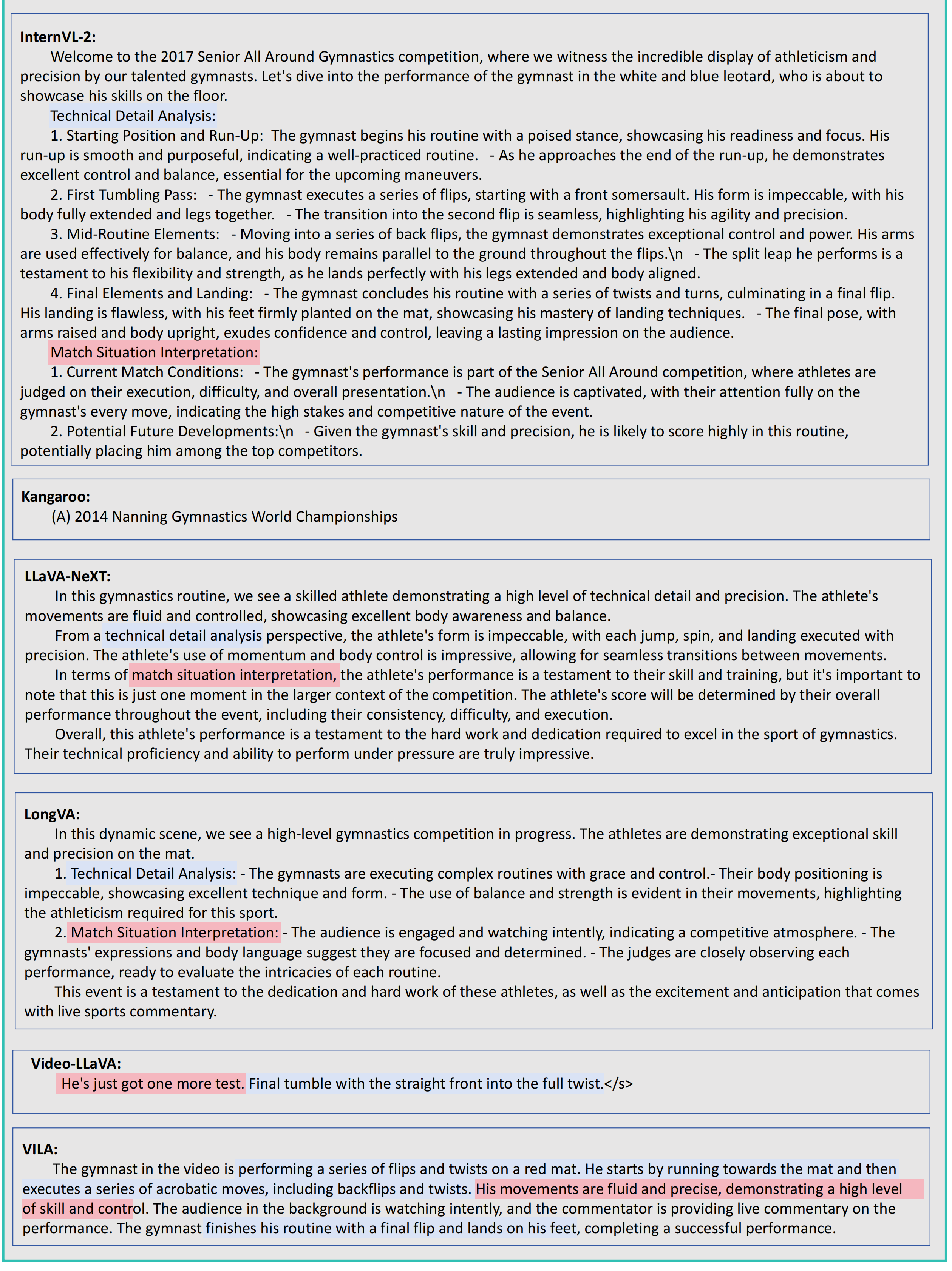}
    \caption{\textbf{An example of the comparison of the generated commentaries in tabletennis.}}
    \label{fig:sample_gym_}
\end{figure*}

%% file: main.bbl
\begin{thebibliography}{46}
\providecommand{\natexlab}[1]{#1}
\providecommand{\url}[1]{\texttt{#1}}
\expandafter\ifx\csname urlstyle\endcsname\relax
  \providecommand{\doi}[1]{doi: #1}\else
  \providecommand{\doi}{doi: \begingroup \urlstyle{rm}\Url}\fi

\bibitem[Anderson et~al.(2016)Anderson, Fernando, Johnson, and Gould]{SPICE}
Peter Anderson, Basura Fernando, Mark Johnson, and Stephen Gould.
\newblock Spice: Semantic propositional image caption evaluation, 2016.

\bibitem[Banerjee and Lavie(2005)]{METEOR}
Satanjeev Banerjee and Alon Lavie.
\newblock {METEOR}: An automatic metric for {MT} evaluation with improved correlation with human judgments.
\newblock In \emph{Proceedings of the {ACL} Workshop on Intrinsic and Extrinsic Evaluation Measures for Machine Translation and/or Summarization}, pages 65--72, Ann Arbor, Michigan, 2005. Association for Computational Linguistics.

\bibitem[Bian et~al.(2024)Bian, Li, Wang, Wang, Huang, Liu, Zhao, Lu, Dou, and Xiong]{bian2024p2anetdatasetbenchmarkdense}
Jiang Bian, Xuhong Li, Tao Wang, Qingzhong Wang, Jun Huang, Chen Liu, Jun Zhao, Feixiang Lu, Dejing Dou, and Haoyi Xiong.
\newblock P2anet: A dataset and benchmark for dense action detection from table tennis match broadcasting videos, 2024.

\bibitem[Brown et~al.(2020)Brown, Mann, Ryder, Subbiah, Kaplan, Dhariwal, Neelakantan, Shyam, Sastry, Askell, Agarwal, Herbert-Voss, Krueger, Henighan, Child, Ramesh, Ziegler, Wu, Winter, Hesse, Chen, Sigler, Litwin, Gray, Chess, Clark, Berner, McCandlish, Radford, Sutskever, and Amodei]{brown2020languagemodelsfewshotlearners}
Tom~B. Brown, Benjamin Mann, Nick Ryder, Melanie Subbiah, Jared Kaplan, Prafulla Dhariwal, Arvind Neelakantan, Pranav Shyam, Girish Sastry, Amanda Askell, Sandhini Agarwal, Ariel Herbert-Voss, Gretchen Krueger, Tom Henighan, Rewon Child, Aditya Ramesh, Daniel~M. Ziegler, Jeffrey Wu, Clemens Winter, Christopher Hesse, Mark Chen, Eric Sigler, Mateusz Litwin, Scott Gray, Benjamin Chess, Jack Clark, Christopher Berner, Sam McCandlish, Alec Radford, Ilya Sutskever, and Dario Amodei.
\newblock Language models are few-shot learners, 2020.

\bibitem[Chen et~al.(2024{\natexlab{a}})Chen, Lin, Zhang, and Huang]{chen2024autoevalvideoautomaticbenchmarkassessing}
Xiuyuan Chen, Yuan Lin, Yuchen Zhang, and Weiran Huang.
\newblock Autoeval-video: An automatic benchmark for assessing large vision language models in open-ended video question answering, 2024{\natexlab{a}}.

\bibitem[Chen et~al.(2024{\natexlab{b}})Chen, Wu, Wang, Su, Chen, Xing, Zhong, Zhang, Zhu, Lu, Li, Luo, Lu, Qiao, and Dai]{chen2024internvlscalingvisionfoundation}
Zhe Chen, Jiannan Wu, Wenhai Wang, Weijie Su, Guo Chen, Sen Xing, Muyan Zhong, Qinglong Zhang, Xizhou Zhu, Lewei Lu, Bin Li, Ping Luo, Tong Lu, Yu Qiao, and Jifeng Dai.
\newblock Internvl: Scaling up vision foundation models and aligning for generic visual-linguistic tasks, 2024{\natexlab{b}}.

\bibitem[Comandur(2022)]{comandur2022sportsreidimprovingreidentification}
Bharath Comandur.
\newblock Sports re-id: Improving re-identification of players in broadcast videos of team sports, 2022.

\bibitem[Deliège et~al.(2021)Deliège, Cioppa, Giancola, Seikavandi, Dueholm, Nasrollahi, Ghanem, Moeslund, and Droogenbroeck]{deliège2021soccernetv2datasetbenchmarksholistic}
Adrien Deliège, Anthony Cioppa, Silvio Giancola, Meisam~J. Seikavandi, Jacob~V. Dueholm, Kamal Nasrollahi, Bernard Ghanem, Thomas~B. Moeslund, and Marc~Van Droogenbroeck.
\newblock Soccernet-v2: A dataset and benchmarks for holistic understanding of broadcast soccer videos, 2021.

\bibitem[Fang et~al.(2024)Fang, Mao, Duan, Zhao, Li, Lin, and Chen]{fang2024mmbenchvideolongformmultishotbenchmark}
Xinyu Fang, Kangrui Mao, Haodong Duan, Xiangyu Zhao, Yining Li, Dahua Lin, and Kai Chen.
\newblock Mmbench-video: A long-form multi-shot benchmark for holistic video understanding, 2024.

\bibitem[Faulkner and Dick(2017)]{Faulkner2017TenniSetAD}
Hayden Faulkner and Anthony~R. Dick.
\newblock Tenniset: A dataset for dense fine-grained event recognition, localisation and description.
\newblock \emph{2017 International Conference on Digital Image Computing: Techniques and Applications (DICTA)}, pages 1--8, 2017.

\bibitem[Feng et~al.(2024)Feng, Lin, Dwivedi, Sun, Patel, and Black]{feng2024chatpose}
Yao Feng, Jing Lin, Sai~Kumar Dwivedi, Yu Sun, Priyanka Patel, and Michael~J Black.
\newblock Chatpose: Chatting about 3d human pose.
\newblock In \emph{Proceedings of the IEEE/CVF Conference on Computer Vision and Pattern Recognition}, pages 2093--2103, 2024.

\bibitem[Fu et~al.(2024)Fu, Dai, Luo, Li, Ren, Zhang, Wang, Zhou, Shen, Zhang, Chen, Li, Lin, Zhao, Li, Xu, Zheng, Chen, Ji, and Sun]{fu2024videommefirstevercomprehensiveevaluation}
Chaoyou Fu, Yuhan Dai, Yongdong Luo, Lei Li, Shuhuai Ren, Renrui Zhang, Zihan Wang, Chenyu Zhou, Yunhang Shen, Mengdan Zhang, Peixian Chen, Yanwei Li, Shaohui Lin, Sirui Zhao, Ke Li, Tong Xu, Xiawu Zheng, Enhong Chen, Rongrong Ji, and Xing Sun.
\newblock Video-mme: The first-ever comprehensive evaluation benchmark of multi-modal llms in video analysis, 2024.

\bibitem[Gao et~al.(2024)Gao, Chen, Cui, Ren, Wang, Zhu, Tian, Ye, He, Zhu, Lu, Lu, Qiao, Dai, and Wang]{gao2024miniinternvlflexibletransferpocketmultimodal}
Zhangwei Gao, Zhe Chen, Erfei Cui, Yiming Ren, Weiyun Wang, Jinguo Zhu, Hao Tian, Shenglong Ye, Junjun He, Xizhou Zhu, Lewei Lu, Tong Lu, Yu Qiao, Jifeng Dai, and Wenhai Wang.
\newblock Mini-internvl: A flexible-transfer pocket multimodal model with 5% parameters and 90% performance, 2024.

\bibitem[Jang et~al.(2017)Jang, Song, Yu, Kim, and Kim]{jang2017tgifqaspatiotemporalreasoningvisual}
Yunseok Jang, Yale Song, Youngjae Yu, Youngjin Kim, and Gunhee Kim.
\newblock Tgif-qa: Toward spatio-temporal reasoning in visual question answering, 2017.

\bibitem[Jin et~al.(2024)Jin, Takanobu, Zhang, Cao, and Yuan]{jin2024chatuniviunifiedvisualrepresentation}
Peng Jin, Ryuichi Takanobu, Wancai Zhang, Xiaochun Cao, and Li Yuan.
\newblock Chat-univi: Unified visual representation empowers large language models with image and video understanding, 2024.

\bibitem[Lewis et~al.(2021)Lewis, Perez, Piktus, Petroni, Karpukhin, Goyal, Küttler, Lewis, tau Yih, Rocktäschel, Riedel, and Kiela]{lewis2021retrievalaugmentedgenerationknowledgeintensivenlp}
Patrick Lewis, Ethan Perez, Aleksandra Piktus, Fabio Petroni, Vladimir Karpukhin, Naman Goyal, Heinrich Küttler, Mike Lewis, Wen tau Yih, Tim Rocktäschel, Sebastian Riedel, and Douwe Kiela.
\newblock Retrieval-augmented generation for knowledge-intensive nlp tasks, 2021.

\bibitem[Li et~al.(2024)Li, Wang, He, Li, Wang, Liu, Wang, Xu, Chen, Luo, Wang, and Qiao]{li2024mvbenchcomprehensivemultimodalvideo}
Kunchang Li, Yali Wang, Yinan He, Yizhuo Li, Yi Wang, Yi Liu, Zun Wang, Jilan Xu, Guo Chen, Ping Luo, Limin Wang, and Yu Qiao.
\newblock Mvbench: A comprehensive multi-modal video understanding benchmark, 2024.

\bibitem[Lin et~al.(2023)Lin, Ye, Zhu, Cui, Ning, Jin, and Yuan]{lin2023videollavalearningunitedvisual}
Bin Lin, Yang Ye, Bin Zhu, Jiaxi Cui, Munan Ning, Peng Jin, and Li Yuan.
\newblock Video-llava: Learning united visual representation by alignment before projection, 2023.

\bibitem[Lin(2004)]{lin-2004-rouge}
Chin-Yew Lin.
\newblock {ROUGE}: A package for automatic evaluation of summaries.
\newblock In \emph{Text Summarization Branches Out}, pages 74--81, Barcelona, Spain, 2004. Association for Computational Linguistics.

\bibitem[Lin et~al.(2024)Lin, Yin, Ping, Lu, Molchanov, Tao, Mao, Kautz, Shoeybi, and Han]{lin2024vilapretrainingvisuallanguage}
Ji Lin, Hongxu Yin, Wei Ping, Yao Lu, Pavlo Molchanov, Andrew Tao, Huizi Mao, Jan Kautz, Mohammad Shoeybi, and Song Han.
\newblock Vila: On pre-training for visual language models, 2024.

\bibitem[Liu et~al.(2024{\natexlab{a}})Liu, Li, Li, Li, Zhang, Shen, and Lee]{liu2024llavanext}
Haotian Liu, Chunyuan Li, Yuheng Li, Bo Li, Yuanhan Zhang, Sheng Shen, and Yong~Jae Lee.
\newblock Llava-next: Improved reasoning, ocr, and world knowledge, 2024{\natexlab{a}}.

\bibitem[Liu et~al.(2024{\natexlab{b}})Liu, Wang, Ma, Wu, Ma, Wei, Jiao, Wu, and Hu]{liu2024kangaroopowerfulvideolanguagemodel}
Jiajun Liu, Yibing Wang, Hanghang Ma, Xiaoping Wu, Xiaoqi Ma, Xiaoming Wei, Jianbin Jiao, Enhua Wu, and Jie Hu.
\newblock Kangaroo: A powerful video-language model supporting long-context video input, 2024{\natexlab{b}}.

\bibitem[Liu et~al.(2022)Liu, Wang, Wang, Ma, and Qiao]{liu2022fineaction}
Yi Liu, Limin Wang, Yali Wang, Xiao Ma, and Yu Qiao.
\newblock Fineaction: A fine-grained video dataset for temporal action localization.
\newblock \emph{IEEE transactions on image processing}, 31:\penalty0 6937--6950, 2022.

\bibitem[Liu et~al.(2024{\natexlab{c}})Liu, Li, Liu, Wang, Ren, Li, Chen, Sun, and Hou]{liu2024tempcompassvideollmsreally}
Yuanxin Liu, Shicheng Li, Yi Liu, Yuxiang Wang, Shuhuai Ren, Lei Li, Sishuo Chen, Xu Sun, and Lu Hou.
\newblock Tempcompass: Do video llms really understand videos?, 2024{\natexlab{c}}.

\bibitem[Liu et~al.(2024{\natexlab{d}})Liu, Xia, Guo, Sun, Shao, and Xia]{liu2024crossblockfinegrainedsemanticcascade}
Zhendong Liu, Haifeng Xia, Tong Guo, Libo Sun, Ming Shao, and Siyu Xia.
\newblock Cross-block fine-grained semantic cascade for skeleton-based sports action recognition, 2024{\natexlab{d}}.

\bibitem[Mangalam et~al.(2023)Mangalam, Akshulakov, and Malik]{mangalam2023egoschemadiagnosticbenchmarklongform}
Karttikeya Mangalam, Raiymbek Akshulakov, and Jitendra Malik.
\newblock Egoschema: A diagnostic benchmark for very long-form video language understanding, 2023.

\bibitem[Miech et~al.(2019)Miech, Zhukov, Alayrac, Tapaswi, Laptev, and Sivic]{miech2019howto100m}
Antoine Miech, Dimitri Zhukov, Jean-Baptiste Alayrac, Makarand Tapaswi, Ivan Laptev, and Josef Sivic.
\newblock Howto100m: Learning a text-video embedding by watching hundred million narrated video clips.
\newblock In \emph{Proceedings of the IEEE/CVF international conference on computer vision}, pages 2630--2640, 2019.

\bibitem[Ning et~al.(2023{\natexlab{a}})Ning, Zhu, Xie, Lin, Cui, Yuan, Chen, and Yuan]{ning2023video}
Munan Ning, Bin Zhu, Yujia Xie, Bin Lin, Jiaxi Cui, Lu Yuan, Dongdong Chen, and Li Yuan.
\newblock Video-bench: A comprehensive benchmark and toolkit for evaluating video-based large language models.
\newblock \emph{arXiv preprint arXiv:2311.16103}, 2023{\natexlab{a}}.

\bibitem[Ning et~al.(2023{\natexlab{b}})Ning, Zhu, Xie, Lin, Cui, Yuan, Chen, and Yuan]{ning2023videobenchcomprehensivebenchmarktoolkit}
Munan Ning, Bin Zhu, Yujia Xie, Bin Lin, Jiaxi Cui, Lu Yuan, Dongdong Chen, and Li Yuan.
\newblock Video-bench: A comprehensive benchmark and toolkit for evaluating video-based large language models, 2023{\natexlab{b}}.

\bibitem[Papineni et~al.(2002)Papineni, Roukos, Ward, and Zhu]{papineni-etal-2002-bleu}
Kishore Papineni, Salim Roukos, Todd Ward, and Wei-Jing Zhu.
\newblock {B}leu: a method for automatic evaluation of machine translation.
\newblock In \emph{Proceedings of the 40th Annual Meeting of the Association for Computational Linguistics}, pages 311--318, Philadelphia, Pennsylvania, USA, 2002. Association for Computational Linguistics.

\bibitem[Pătrăucean et~al.(2023)Pătrăucean, Smaira, Gupta, Continente, Markeeva, Banarse, Koppula, Heyward, Malinowski, Yang, Doersch, Matejovicova, Sulsky, Miech, Frechette, Klimczak, Koster, Zhang, Winkler, Aytar, Osindero, Damen, Zisserman, and Carreira]{pătrăucean2023perceptiontestdiagnosticbenchmark}
Viorica Pătrăucean, Lucas Smaira, Ankush Gupta, Adrià~Recasens Continente, Larisa Markeeva, Dylan Banarse, Skanda Koppula, Joseph Heyward, Mateusz Malinowski, Yi Yang, Carl Doersch, Tatiana Matejovicova, Yury Sulsky, Antoine Miech, Alex Frechette, Hanna Klimczak, Raphael Koster, Junlin Zhang, Stephanie Winkler, Yusuf Aytar, Simon Osindero, Dima Damen, Andrew Zisserman, and João Carreira.
\newblock Perception test: A diagnostic benchmark for multimodal video models, 2023.

\bibitem[Sanders and Van~Durme(2024)]{sanders2024survey}
Kate Sanders and Benjamin Van~Durme.
\newblock A survey of video datasets for grounded event understanding.
\newblock In \emph{Proceedings of the IEEE/CVF Conference on Computer Vision and Pattern Recognition}, pages 7314--7327, 2024.

\bibitem[Shao et~al.(2020)Shao, Zhao, Dai, and Lin]{shao2020finegymhierarchicalvideodataset}
Dian Shao, Yue Zhao, Bo Dai, and Dahua Lin.
\newblock Finegym: A hierarchical video dataset for fine-grained action understanding, 2020.

\bibitem[Shao et~al.(2024)Shao, Qian, Xiao, Song, Zong, Wang, Liu, and Li]{shao2024visualcotadvancingmultimodal}
Hao Shao, Shengju Qian, Han Xiao, Guanglu Song, Zhuofan Zong, Letian Wang, Yu Liu, and Hongsheng Li.
\newblock Visual cot: Advancing multi-modal language models with a comprehensive dataset and benchmark for chain-of-thought reasoning, 2024.

\bibitem[Solberg et~al.(2024)Solberg, Sarkhoosh, Gautam, Sabet, Halvorsen, and Midoglu]{solberg2024playertvadvancedplayertracking}
Håkon~Maric Solberg, Mehdi~Houshmand Sarkhoosh, Sushant Gautam, Saeed~Shafiee Sabet, Pål Halvorsen, and Cise Midoglu.
\newblock Playertv: Advanced player tracking and identification for automatic soccer highlight clips, 2024.

\bibitem[Team()]{InternVL2}
OpenGVLab Team.
\newblock Internvl2: Better than the best—expanding performance boundaries of open-source multimodal models with the progressive scaling strategy.
\newblock Accessed: 2024-11-14.

\bibitem[Vedantam et~al.(2014)Vedantam, Zitnick, and Parikh]{cider}
Ramakrishna Vedantam, C.~Lawrence Zitnick, and Devi Parikh.
\newblock Cider: Consensus-based image description evaluation.
\newblock \emph{CoRR}, abs/1411.5726, 2014.

\bibitem[Wang et~al.(2024)Wang, He, Hong, Cheng, Zhang, Qi, Huang, Xu, Dong, Ding, and Tang]{wang2024lvbenchextremelongvideo}
Weihan Wang, Zehai He, Wenyi Hong, Yean Cheng, Xiaohan Zhang, Ji Qi, Shiyu Huang, Bin Xu, Yuxiao Dong, Ming Ding, and Jie Tang.
\newblock Lvbench: An extreme long video understanding benchmark, 2024.

\bibitem[Wei et~al.(2023)Wei, Wang, Schuurmans, Bosma, Ichter, Xia, Chi, Le, and Zhou]{wei2023chainofthoughtpromptingelicitsreasoning}
Jason Wei, Xuezhi Wang, Dale Schuurmans, Maarten Bosma, Brian Ichter, Fei Xia, Ed Chi, Quoc Le, and Denny Zhou.
\newblock Chain-of-thought prompting elicits reasoning in large language models, 2023.

\bibitem[Wu et~al.(2024)Wu, He, Wu, and Wang]{wu2024sportshhidatasethumanhumaninteraction}
Tao Wu, Runyu He, Gangshan Wu, and Limin Wang.
\newblock Sportshhi: A dataset for human-human interaction detection in sports videos, 2024.

\bibitem[Xiao et~al.(2021)Xiao, Shang, Yao, and Chua]{2021NExT}
Junbin Xiao, Xindi Shang, Angela Yao, and Tat~Seng Chua.
\newblock Next-qa:next phase of question-answering to explaining temporal actions.
\newblock 2021.

\bibitem[Xu et~al.(2017)Xu, Zhao, Xiao, Wu, Zhang, He, and Zhuang]{2017Video}
Dejing Xu, Zhou Zhao, Jun Xiao, Fei Wu, Hanwang Zhang, Xiangnan He, and Yueting Zhuang.
\newblock Video question answering via gradually refined attention over appearance and motion.
\newblock In \emph{ACM}, pages 1645--1653, 2017.

\bibitem[Xu et~al.(2023)Xu, Ye, Wu, Yan, Miao, Ye, Xu, Hu, Shi, Xu, Li, Qian, Que, Zhang, Zeng, and Huang]{xu2023youkumplug10millionlargescale}
Haiyang Xu, Qinghao Ye, Xuan Wu, Ming Yan, Yuan Miao, Jiabo Ye, Guohai Xu, Anwen Hu, Yaya Shi, Guangwei Xu, Chenliang Li, Qi Qian, Maofei Que, Ji Zhang, Xiao Zeng, and Fei Huang.
\newblock Youku-mplug: A 10 million large-scale chinese video-language dataset for pre-training and benchmarks, 2023.

\bibitem[Yu et~al.(2019)Yu, Xu, Yu, Yu, Zhao, Zhuang, and Tao]{yu2019activitynetqadatasetunderstandingcomplex}
Zhou Yu, Dejing Xu, Jun Yu, Ting Yu, Zhou Zhao, Yueting Zhuang, and Dacheng Tao.
\newblock Activitynet-qa: A dataset for understanding complex web videos via question answering, 2019.

\bibitem[Zhang et~al.(2024)Zhang, Zhang, Li, Zeng, Yang, Zhang, Wang, Tan, Li, and Liu]{zhang2024longcontexttransferlanguage}
Peiyuan Zhang, Kaichen Zhang, Bo Li, Guangtao Zeng, Jingkang Yang, Yuanhan Zhang, Ziyue Wang, Haoran Tan, Chunyuan Li, and Ziwei Liu.
\newblock Long context transfer from language to vision, 2024.

\bibitem[Zhou et~al.(2024)Zhou, Wang, Liu, Hao, Hui, Tarkoma, and Kangasharju]{zhou2024survey}
Pengyuan Zhou, Lin Wang, Zhi Liu, Yanbin Hao, Pan Hui, Sasu Tarkoma, and Jussi Kangasharju.
\newblock A survey on generative ai and llm for video generation, understanding, and streaming.
\newblock \emph{arXiv preprint arXiv:2404.16038}, 2024.

\end{thebibliography}
